\documentclass[10pt,journal,compsoc]{IEEEtran}
\newif\ifpreprint
\preprinttrue
\newif\ifsupplement
\supplementfalse

\usepackage{amsmath,amsfonts,bm}

\def\eqref#1{equation~\ref{#1}}

\def\1{\bm{1}}

\def\vmu{{\bm{\mu}}}

\def\vepsilon{{\bm{\epsilon}}}
\def\va{{\bm{a}}}

\def\vc{{\bm{c}}}

\def\vh{{\bm{h}}}

\def\vn{{\bm{n}}}

\def\vt{{\bm{t}}}
\def\vu{{\bm{u}}}
\def\vv{{\bm{v}}}
\def\vw{{\bm{w}}}
\def\vx{{\bm{x}}}

\def\vz{{\bm{z}}}

\def\mC{{\bm{C}}}

\def\mH{{\bm{H}}}
\def\mI{{\bm{I}}}

\def\mS{{\bm{S}}}

\def\mU{{\bm{U}}}
\def\mV{{\bm{V}}}

\def\mZ{{\bm{Z}}}

\def\mSigma{{\bm{\Sigma}}}
\def\mOmega{{\bm{\Omega}}}

\DeclareMathAlphabet{\mathsfit}{\encodingdefault}{\sfdefault}{m}{sl}
\SetMathAlphabet{\mathsfit}{bold}{\encodingdefault}{\sfdefault}{bx}{n}

\def\gD{{\mathcal{D}}}

\DeclareMathOperator*{\argmin}{arg\,min}

\usepackage{microtype}
\usepackage{graphicx}
\usepackage{subcaption}
\usepackage{booktabs} 
\usepackage[table,dvipsnames]{xcolor}
\usepackage{multirow}
\usepackage{enumitem}
\usepackage{pifont}
\usepackage{algorithm}
\usepackage{algorithmic}
\ifpreprint
  \newcommand{\appref}[1]{\ref{#1}}
  \newcommand{\mainref}[1]{\ref{#1}}
\else
  \usepackage{xr-hyper}
  \ifsupplement
    \newcommand{\mainref}[1]{\ref{M:#1}}
    \newcommand{\appref}[1]{\ref{#1}}
  \else
    \newcommand{\appref}[1]{\ref{S:#1}}
    \newcommand{\mainref}[1]{\ref{#1}}
  \fi
\fi
\usepackage{hyperref}
\usepackage{url}
\usepackage{orcidlink}

\newcommand{\para}[1]{\smallskip\noindent\textbf{#1}\hspace{0.45em}}

\begin{document}
\bstctlcite{IEEEexample:BSTcontrol}
\title{Towards Fast and Disentangled Counterfactuals for Visual Foundation Models}

\author{Sidney~Bender~\orcidlink{0009-0007-1727-8577},
        Benedikt~Kunz~\orcidlink{0009-0006-9404-6924},
        Ahmed~Zeid~\orcidlink{0009-0001-4375-6356},
        Shinichi~Nakajima~\orcidlink{0000-0003-3970-4569},
        Klaus-Robert~M\"uller~\orcidlink{0000-0002-3861-7685},
        Marco~Morik~\orcidlink{0009-0001-7897-2243}
\IEEEcompsocitemizethanks{
    \IEEEcompsocthanksitem S. Bender, K.-R. M\"uller, S. Nakajima, and M. Morik are also with the Berlin Institute for the Foundations of Learning and Data (BIFOLD), Berlin, Germany.
    \IEEEcompsocthanksitem S. Bender, B. Kunz, A. Zeid, S. Nakajima, K.-R. M\"uller, and M. Morik are with the Machine Learning Group, Technische Universit\"at Berlin, Berlin, Germany.\protect\\
    E-mail: \{nakajima, klaus-robert.mueller, m.morik\}@tu-berlin.de
}%
}

\IEEEtitleabstractindextext{%
\begin{abstract}
Foundation models remain vulnerable to spurious correlations and ``Clever Hans'' strategies. Explainable machine learning can find and remove such strategies for classifiers without metadata. For foundation models, no such option exists yet.
We propose Disentangled Diffusion Autoencoders (DiDAE). DiDAE wraps a frozen foundation model in a conditional diffusion decoder. A counterfactual is one closed-form edit along a direction of a disentangled dictionary, followed by decoding. The dictionary can be supervised (Procrustes) or unsupervised (Singular Value Decomposition, Sparse Autoencoders).
No gradients are needed, so DiDAE is up to 2000 times faster than the state of the art. We evaluate on six datasets, two synthetic and four real-world. In a desiderata-driven benchmark on three of them, its counterfactuals are on par with or better than the state of the art, and they repair downstream classifiers through Counterfactual Knowledge Distillation (CFKD), where they beat metadata-based correction.
The same machinery can rank a pretrained dictionary against a trained classifier. It returns the few directions the classifier actually reads, each causally verified by a counterfactual that flips the decision, and repairs the classifier along those a teacher marks spurious.
The workflow is plug-and-play in our open-source \textsc{Peal} library we publish alongside the paper. With a public dictionary and a pretrained decoder, all that remains is a cheap linear distillation of the classifier and its own fine-tuning.
\end{abstract}

}

\maketitle

\IEEEdisplaynontitleabstractindextext
\IEEEpeerreviewmaketitle

\begin{figure*}[!t]
\centering
\includegraphics[width=\textwidth]{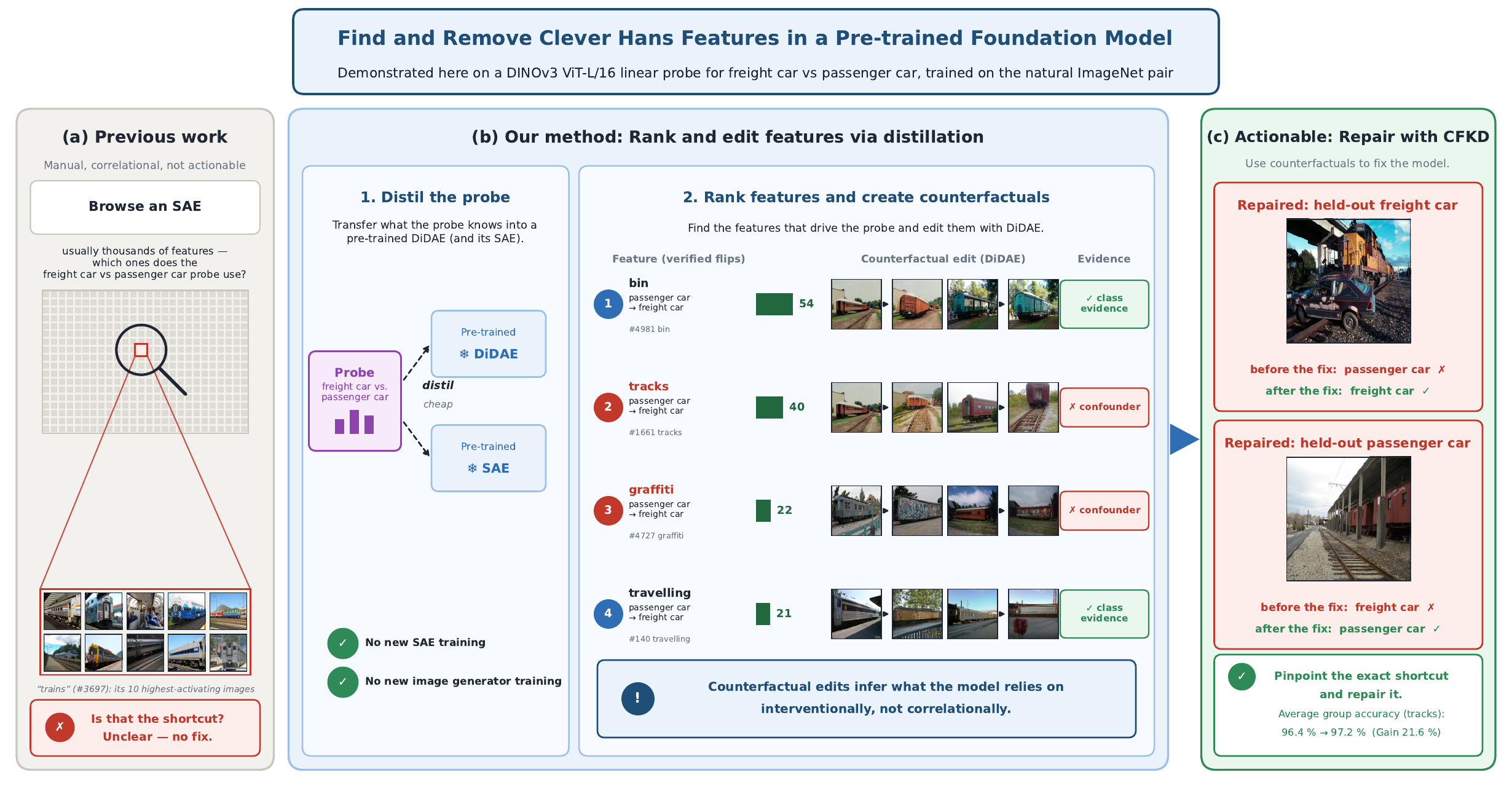}
\caption{
\textbf{Find and remove a Clever Hans feature of a classifier without training a generator or a dictionary.}
The classifier is a DINOv3~\cite{simeoni2025dinov3} ViT-L/16 linear head trained on the natural ImageNet \emph{freight car} vs.\ \emph{passenger car} pair; the dictionary is the public $6{,}144$-atom MSAE~\cite{zaigrajew2025interpreting} over OpenAI CLIP ViT-L/14 and the decoder our pretrained ImageNet RAE (App.~\appref{app:rae}), both taken as they are. \textbf{(a)} The usual route, browsing the dictionary by highest-activating images, returns the atom that dominates this dataset, \texttt{\#3697} (``trains''), which says nothing about the decision. \textbf{(b)} DiDAE distils the classifier into the encoder space, a cheap closed-form step, and ranks the atoms by counterfactuals that flip the real classifier.
Behind the class evidence ``bin'' rank rail tracks ($40$ verified flips) and graffiti, the one spurious feature Neuhaus et al.~\cite{neuhaus2023spurious} report for freight car; the tracks shortcut is new. Each direction comes as before/after pairs (two chosen per direction here), so a practitioner sees what it means and marks it as class evidence or confounder. The pairs also correct the atoms' automatic CLIP-Dissect~\cite{oikarinen2023clipdissect} names: ``bin'' and ``travelling'' change the car body and are class evidence, which the words do not suggest, and ``tracks'' is not rails anywhere in the image but a central view down a receding track.
\textbf{(c)} One CFKD iteration on the tracks direction raises average group accuracy over the class$\times$tracks groups from $96.4\%$ to $97.2\%$, a Gain of $21.6\%$ on an already accurate probe, without touching the backbone.
}
\label{fig:teaser}
\end{figure*}

\section{Introduction}
\label{section:intro}

Deep learning models, despite impressive performance on benchmarks, remain highly vulnerable to spurious correlations, often adopting Clever Hans strategies that fail to generalize out-of-distribution~\cite{lapuschkin2019unmasking, geirhos2020shortcut}. While foundation models (FMs) like CLIP~\cite{radford2021learning} have demonstrated robust few-shot capabilities, recent studies indicate they systematically encode non-causal artifacts such as background textures~\cite{kauffmann2025explainable} or technical features~\cite{komen2025towards}.

Current mitigation strategies typically rely on explicit group labels to reweight underrepresented subgroups (e.g., GroupDRO~\cite{sagawa2020distributionally}). These methods do not work when labels are unavailable or confounding variables are unknown. Explainable AI offers an alternative via \textit{Counterfactual Knowledge Distillation} (CFKD)~\cite{bender2023towards}, which generates counterfactuals to expose and prune reliance on confounders. However, the efficacy of CFKD is bottlenecked by the quality and speed of Visual Counterfactual Explainers (VCEs).

\begin{figure*}[t]
\centering
\includegraphics[scale=.2]{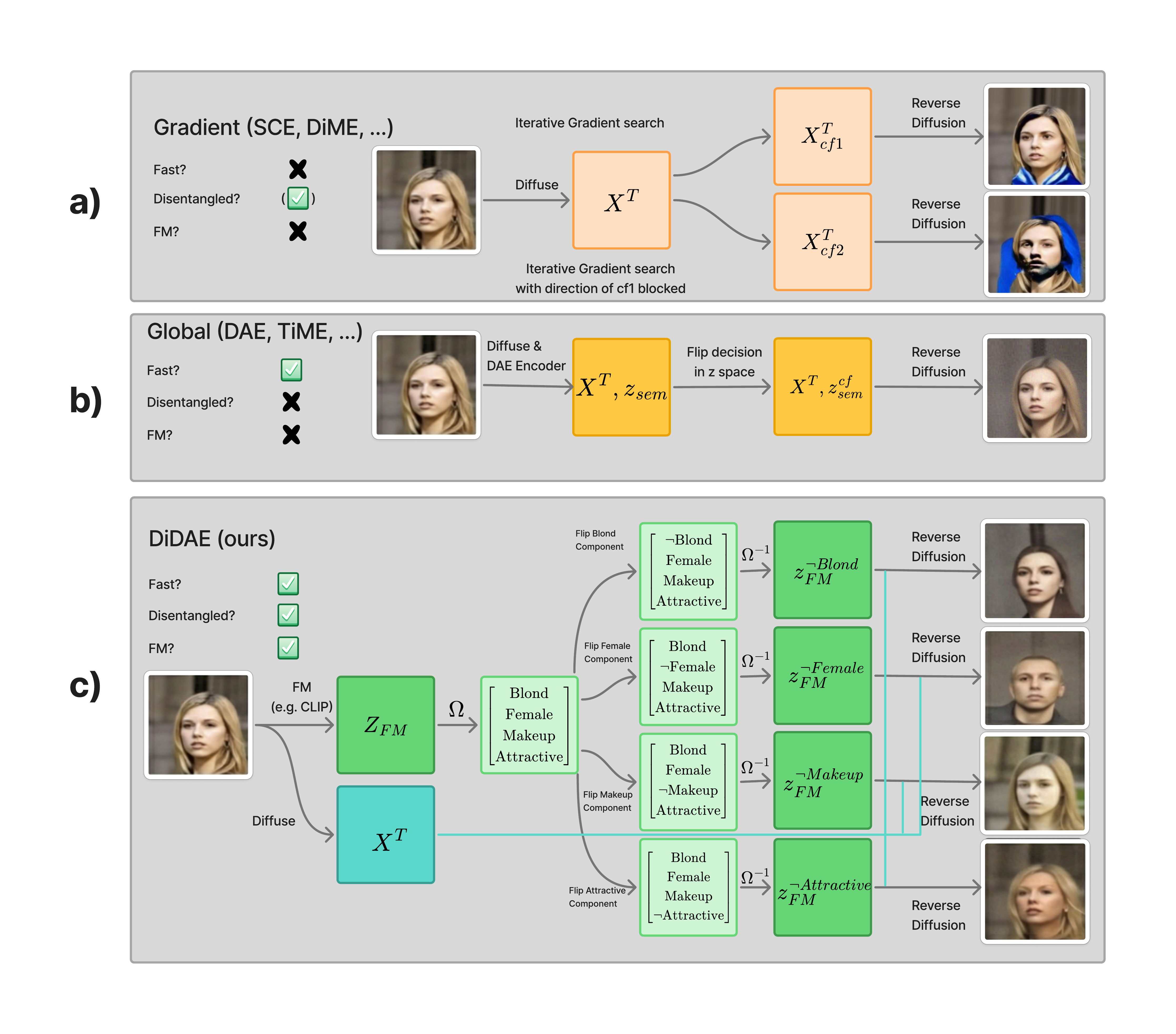}
\caption{Comparison of traditional gradient-based counterfactuals (a), global counterfactual methods (b) versus the proposed DiDAE approach (c) on a CelebA classifier trained on the ``Blond Hair'' label. The label is spuriously correlated with ``Female'', ``Heavy Makeup'' and ``Attractive''. Gradient-based methods require slow, iterative gradient updates through the diffusion process; global counterfactual methods like diffusion autoencoders or TIME are fast, but only produce one counterfactual per factual often either too weak to be seen clearly (as in the figure) or with entangled changes (e.g., changing hair color and the correlated gender simultaneously). Both are only able to explain classifiers and can not explain the foundation models themselves. In contrast, DiDAE utilizes a frozen foundation model to decompose embeddings $\vz_{\mathrm{FM}}$ into disentangled semantic components. Counterfactuals are generated via a single closed-form linear edit in this semantic space, which moves one component to the bound of its empirical range, followed by decoding via a diffusion decoder. This is fast, can create diverse, disentangled counterfactuals, and can be applied directly to the foundation model representations as well, without the need to explain a specific classifier.}
\label{fig:overview}
\end{figure*}

As illustrated in Figure~\ref{fig:overview}a, state-of-the-art VCEs like SCE~\cite{bender2025desiderata} rely on iterative gradient-based optimization. This process is generally slow, often yields adversarial noise rather than semantic changes, and creates entangled edits. While gradient-free methods have been proposed to improve generation speed~\cite{jeanneret2024text,sobieski2024global, cao2025leapfactual}, they typically lack mechanisms for explicit semantic sparsification and diversification, limiting their utility for precise model correction.

To address these limitations, we introduce \textbf{Disentangled Diffusion Autoencoders (DiDAE)}\footnote{A preliminary, non-archival version of this work was presented at the ICLR 2026 Trustworthy AI workshop~\cite{bender2026visual}. This paper presents the fully extended framework.}.
As shown in Figure~\ref{fig:overview}c, our DiDAE framework combines frozen foundation models with disentangled dictionary learning. By moving a sample along one learned semantic component at a time, as far as the data support, DiDAE generates disentangled, diverse counterfactuals without gradient updates.

Our main contributions are as follows:

\begin{itemize}
    \item \textbf{DiDAE}, a diffusion autoencoder on frozen foundation-model representations that
    generates counterfactuals by gradient-free, component-wise edits in a disentangled dictionary,
    supervised (Procrustes) or unsupervised (SVD, SAE~\cite{bricken2023towards}).

    \item \textbf{Clever Hans mitigation} with these counterfactuals through
    CFKD~\cite{bender2023towards}, where a pre-clustered teacher needs one decision per direction.

    \item \textbf{Dictionary ranking}: the same machinery orders a pretrained dictionary by
    \emph{verified ambient flips}, decoded edits that flip the real classifier and demonstrably
    moved the intended component, and repairs the classifier along the directions a teacher
    marks spurious. Only a cheap linear distillation and the classifier's own fine-tuning
    remain (Figure~\ref{fig:teaser})\footnote{DiDAE ships as part of \textsc{Peal}~\cite{bender2026peal}, our open-source PyTorch Explain-and-Adapt library, and runs at the moment of publication out of the box on a user-supplied classifier and dataset in a web interface: \url{https://peal.ml.tu-berlin.de}. The library fetches the weights of our pretrained natural-image decoders and the public dictionaries automatically from Hugging Face, so the workflow of Figure~\ref{fig:teaser} needs no generator training on the user's side.}.
\end{itemize}

We conduct a comprehensive, desiderata-driven evaluation on synthetic, natural, and medical imaging datasets, comparing DiDAE against state-of-the-art visual counterfactual explanation methods and metadata-based Clever Hans mitigation approaches, demonstrating both high-quality visual explanations and the superior capability of DiDAE-generated counterfactuals to improve the robustness of downstream classifiers at up to 2000 times the speed. Section~\ref{section:methods} presents the method in the order the pipeline is built, Section~\ref{sec:experiments} the datasets, instances and protocols we evaluate, and Section~\ref{section:results} the results.

\begin{figure*}[t]
\centering
\includegraphics[width=\textwidth]{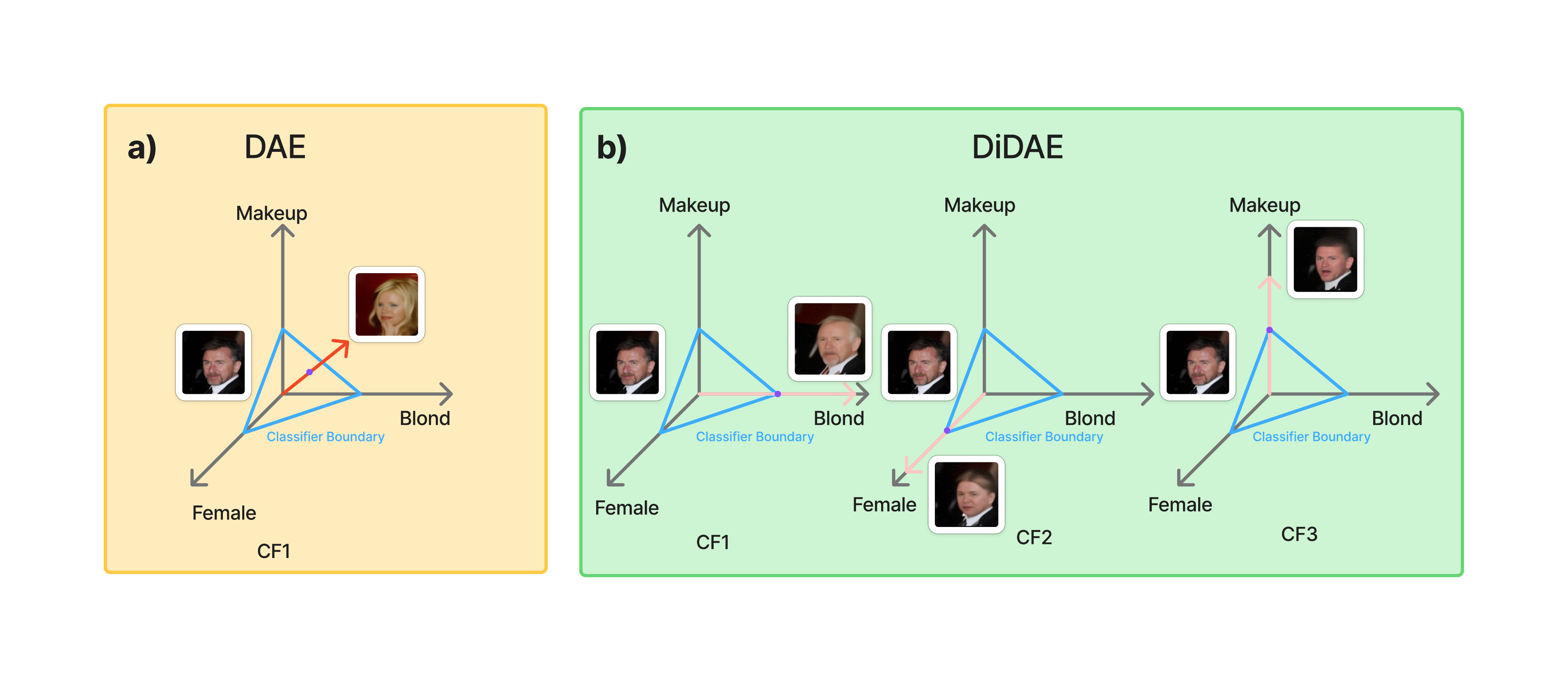}
\caption{\textbf{(a)} Global methods like DAE generate counterfactuals by moving orthogonally to the classifier's decision boundary (red arrow). This trajectory entangles multiple variables (e.g., hair color, makeup, and gender), altering them simultaneously and yielding a less interpretable result. \textbf{(b)} In contrast, DiDAE disentangles these transformations along distinct semantic axes based on a learned dictionary. By moving independently along these axes (pink arrows), DiDAE produces fine-grained, interpretable counterfactuals (CF1, CF2, CF3) that isolate attribute-specific changes across the classifier boundary.}
\label{fig:dae_vs_didae}
\end{figure*}

\section{Related Work}
\label{section:related}

There are three corpora of related work relevant for our work: the correction of models relying on spurious correlations, visual counterfactual explainers, and interpretability of foundation models. For a map of the explainable-AI methods these corpora draw on we refer to the overview of~\cite{holzinger2022explainable}; the XAI 2.0 manifesto~\cite{longo2024explainable} lists the open problems the field has set itself, and three of them, augmenting attribution methods, evaluating explanations, and turning an explanation into an improved model, are the ones the three corpora below and this paper address.

\para{Spurious Correlations and Model Correction.}
Standard approaches to mitigate spurious correlations assume access to confounder annotations.
Projection~\cite{bolukbasi2016man, chuang2023debiasing} linearly removes specific component directions from the embedding space directly in a foundation model so that a classifier probed or fine-tuned from it can not learn the spurious correlation.
Distributional robustness methods like GroupDRO~\cite{sagawa2020distributionally} and DFR~\cite{kirichenko2023last} optimize worst-group performance of downstream classifiers, but struggle when minority groups are too small or unknown.
P-ClArC~\cite{anders2022finding} and RR-ClArC~\cite{dreyer2024hope, pahde2025navigating} estimate an artifact's direction in the model's latent space from samples annotated with it, and project it out or penalize the model's sensitivity along it; EGEM~\cite{linhardt2024preemptively} prunes Clever Hans strategies preemptively.
Feature attributions like LRP~\cite{bach2015pixel} can be leveraged with DRSA to find~\cite{chormai2024disentangled} and with Subdistill~\cite{chormai2026distilling} to distill spurious correlations~\cite{chormai2026investigating} even more effectively.
However, attribution maps are hard to act on: in user studies they helped people judge a model's decisions no better than nearest training examples, and automatic attribution scores barely predicted that usefulness~\cite{nguyen2021effectiveness}.
CFKD~\cite{bender2023towards, bender2025mitigating, hackstein2025imbalanced, delzer2026reproducibility} addresses this by using counterfactuals for data augmentation, but the generation speed and quality of the underlying counterfactuals has historically bottlenecked its performance.
The teacher of CFKD is an instance of interactive, human-in-the-loop machine learning, which~\cite{holzinger2016interactive} argues is needed exactly when the training data cannot settle a question on their own, as with a confounder that no annotation names; what DiDAE changes is the granularity of that interaction, one decision per direction instead of one per counterfactual (Section~\ref{sec:didae_cfkd}).

\para{Visual Counterfactual Explainers (VCEs).}
While there are counterfactual explainers for other domains like tabular data, natural language~\cite{dehghanighobadi2025can}, graphs~\cite{bechtoldt2025graph, bechtoldt2026faithful} and proteins~\cite{klos2026protein} we will focus on visual counterfactuals in this paper.
Generating valid visual counterfactuals is a challenging inverse problem. While early methods used GANs, VAEs, or Normalizing Flows (e.g., DiVE~\cite{rodriguez2021beyond}, LatentShift~\cite{cohen2021gifsplanation} and Diffeomorphic Counterfactuals~\cite{dombrowski2024diffeomorphic}), the field has shifted rapidly toward diffusion and flow-matching approaches. Proximal on-manifold counterfactuals can now be generated using methods such as DVCE~\cite{augustin2022diffusion}, DiME~\cite{jeanneret2022diffusion}, and Diff-ICE~\cite{pegios2025diffusion}. More specialized approaches like  CDCT~\cite{varshney2024generating} focus on generating counterfactual trajectories for concept discovery or for regression models~\cite{ha2025diffusion}. Recent advancements have also targeted semantic sparsity and computational efficiency. ACE~\cite{jeanneret2023adversarial} and FastDiME~\cite{weng2024fast} introduce mechanisms to generate semantically disentangled edits, while SCE~\cite{bender2025desiderata} and SCE-lite-HQ~\cite{zeid2026sce} explicitly optimize for a diverse set of counterfactuals. However, they are slow due to their need to propagate gradients through the rollouts of diffusion models iteratively, and they cannot explain foundation models.
Moreover, the sparsification and diversification of SCE are constrained to spatially separable features.
If features should overlap, e.g. because one of them is global, so far the mechanisms do not work.
While there exist gradient-free visual counterfactual explainers like the Diffusion Autoencoder~\cite{preechakul2022diffusion}, TIME~\cite{jeanneret2024text}, GCD~\cite{sobieski2024global}, and LeapFactual~\cite{cao2025leapfactual}, which improve generation speed, they lack mechanisms for explicit semantic sparsification and diversification, limiting their utility for precise model analysis and correction (Figure~\ref{fig:dae_vs_didae}).
All of them produce one edit direction per classifier, so in our benchmark DAE stands in for the
family: TIME, GCD and LeapFactual share its single-direction structure and would separate from
DiDAE on the same desideratum, diversity, for the same reason.
The building block DiDAE rests on, a diffusion decoder conditioned on a frozen CLIP image embedding, was introduced by unCLIP~\cite{ramesh2022hierarchical}, which also inverts images with DDIM and moves the CLIP embedding to produce variations and text-guided edits. unCLIP has no dictionary, no trust region, no link to a classifier and no verification, so its edits are neither disentangled nor tied to a decision; DiDAE adds exactly these.
DiDAE is gradient-free and fast and yet capable of semantic sparsification and diversification even of spatially overlapping features.

\para{Interpretability of Foundation Models.}
While there is some recent work explicitly aligning visual foundation models to the visual perception of humans~\cite{muttenthaler2025aligning} and making them inherently more interpretable, and there exist gradient-based explanations for specialized architectures like Mamba~\cite{rezaei2024mambalrp}, recent work has focused on interpreting the latent spaces of FMs using Sparse Autoencoders (SAEs)~\cite{bricken2023towards, bussmann2024batchtopk, bussmann2025learning, zhu2025abstopk, korznikov2025ortsae, bhalla2025temporal, klotz2026evaluating}. These methods decompose dense embeddings into interpretable ``concepts'', but leave two questions open. The first is what a concept learned without supervision actually means: the standard answer, its highest-activating examples, is hard to read when concepts are correlated or entangled. The second is which of the thousands of atoms a given downstream decision relies on: activation galleries are correlational and tied to no classifier, so an atom can top-activate on a spurious cue and be causally inert, or be decisive yet never top any list. DiDAE addresses the first by rendering a sample with a component activated next to the closest sample with it deactivated, and, run in its ranking mode, the second by ordering atoms by the counterfactuals that flip the classifier under study.

\section{Methods}
\label{section:methods}

DiDAE turns a frozen visual foundation model $\Phi$ into an editable diffusion autoencoder. Throughout,
$\Phi$ is used as a frozen encoder that maps an image $\vx$ to a representation
$\vz_{\mathrm{FM}}=\Phi(\vx)\in\mathbb{R}^{D}$. A \emph{dictionary}
$\mOmega=[\vv_1,\ldots,\vv_{D'}]\in\mathbb{R}^{D\times D'}$ supplies $D'$ unit-norm directions
(atoms) in that space, and a \emph{component encoder} $g_{\Omega}$ reads off the coefficients
$\vc=g_{\Omega}(\vz_{\mathrm{FM}})\in\mathbb{R}^{D'}$ of a representation in the dictionary, so that
$\vz_{\mathrm{FM}}\approx\bar{\vz}+\mOmega\vc$ with $\bar{\vz}$ the mean representation. An
intervention on component $k$ replaces its coefficient $c_k$ by a target value $c'_k$ and moves
the representation along the atom,
\begin{equation}
    \vz'_{\mathrm{FM}}=\vz_{\mathrm{FM}}+\delta_k\,\vv_k,
    \qquad
    \delta_k=c'_k-c_k .
    \label{eq:component_intervention}
\end{equation}
A conditional generative decoder then renders $\vz'_{\mathrm{FM}}$ as an image $\tilde{\vx}$ that
keeps everything about $\vx$ the edit did not touch. Figure~\ref{fig:overview}c sketches this flow
for one edit and Figure~\ref{fig:dae_vs_didae}b its geometry, one axis at a time;
Figure~\ref{fig:teaser} shows the ranking-and-repair workflow built on it
(Sections~\ref{sec:didae_assemled} and~\ref{sec:didae_cfkd}).

Four ingredients make this work, and each is a slot rather than a fixed choice: the encoder
$\Phi$, the dictionary $(\mOmega,g_{\Omega})$, the decoder, and the inversion that pairs a given
image with a latent code of the decoder. Two of them are fitted \emph{once, offline, and per
domain rather than per classifier}: the decoder (Section~\ref{sec:conditional_score_field}) and
the dictionary (Section~\ref{sec:disentangled_component_analysis}). Both steps are optional
whenever the domain already provides the parts. For natural images at ImageNet scale a public SAE
over CLIP exists and the matching decoder is downloaded by the library on demand, so a practitioner trains no generative model
and starts at the per-image steps with only the classifier to be explained, which is distilled in closed form
(Section~\ref{subsec:linesearch_factor}); Figure~\ref{fig:teaser} is that case. The remaining
sections are the per-image inference steps: how a direction becomes an edit that stays inside the
data support (Section~\ref{subsec:linesearch_factor}), how that edit is carried back to pixels
(Section~\ref{sec:inversion}), how the pieces assemble into a counterfactual explainer and, given
a classifier, into a ranking of the whole dictionary (Section~\ref{sec:didae_assemled}), and how
the counterfactuals correct a downstream model (Section~\ref{sec:didae_cfkd}). The instances of the
four slots we use on each dataset are listed in Section~\ref{sec:model_architectures}
(Table~\ref{tab:components}).

\subsection{Conditional Decoder}
\label{sec:conditional_score_field}

The decoder is a conditional generative model $p_\theta(\vx\mid\vz_{\mathrm{FM}})$ with parameters
$\theta$, trained on pairs $(\vx,\Phi(\vx))$ while $\Phi$ stays frozen. This is the one departure
from a standard diffusion autoencoder~\cite{preechakul2022diffusion}, which learns its semantic
encoder jointly with the decoder: keeping $\Phi$ fixed means the decoder learns to render exactly
the representation the foundation model already computes, so an edit of $\vz_{\mathrm{FM}}$ is an
edit of what the foundation model sees rather than of a surrogate space. Any conditional diffusion
or flow-matching model~\cite{ho2020denoising,rombach2022high,lipman2022flow} qualifies, provided
the conditioning enters its score network and the model admits an inversion
(Section~\ref{sec:inversion}). For a denoising diffusion model with noise-prediction network
$\epsilon_\theta$, for instance, training draws a diffusion time $t$ and noise
$\vepsilon\sim\mathcal{N}(\mathbf{0},\mI)$, forms the noisy image
$\vx_t=\sqrt{\bar\alpha_t}\,\vx+\sqrt{1-\bar\alpha_t}\,\vepsilon$ with $\bar\alpha_t$ the
cumulative signal coefficient of the noise schedule, and minimizes the familiar objective
\begin{equation}
    \mathcal{L}(\theta)
    =\mathbb{E}_{\vx,t,\vepsilon}
    \Big[\big\|\epsilon_\theta\big(\vx_t,t,\Phi(\vx)\big)-\vepsilon\big\|_2^2\Big].
    \label{eq:diffusion_training_objective}
\end{equation}
The only change to the usual recipe is that the gradient stops at $\theta$ and never reaches
$\Phi$. The three decoders we train, a pixel-space diffusion autoencoder, a conditioned latent
diffusion model and a representation autoencoder whose score network operates on a grid of frozen
patch tokens, differ only in what is denoised and how $\vz_{\mathrm{FM}}$ enters the network;
their architectures are given in Section~\ref{sec:model_architectures} and
Appendices~\appref{app:diffae}, \appref{app:pathldm} and~\appref{app:rae}.

\subsection{Disentangled Dictionaries}
\label{sec:disentangled_component_analysis}

The dictionary decides what ``one component'' means. DiDAE is agnostic to where it comes from:
any set of directions together with an encoder that reads off their coefficients will do,
including hand-specified concept vectors. We use three constructions that differ in how much
supervision they need. Let $\mZ\in\mathbb{R}^{N\times D}$ collect the centered representations
$\Phi(\vx_i)-\bar{\vz}$ of $N$ training images.

\para{Supervised alignment (orthogonal Procrustes).}
When $p$ semantic variables are annotated, collected in $\mS\in\mathbb{R}^{N\times p}$ (known
generative factors or attribute labels), we look for orthonormal directions
$\mOmega_1\in\mathbb{R}^{D\times p}$ whose coefficients track them:
\begin{equation}
    \min_{\mOmega_1}\ \|\mZ\mOmega_1-\mS\|_F^2
    \quad\text{s.t.}\quad
    \mOmega_1^{\top}\mOmega_1=\mI_p .
    \label{eq:procrustes_objective}
\end{equation}
The solution is closed-form: with the thin singular value decomposition of the cross-covariance
$\mZ^{\top}\mS=\mU\mSigma\mV^{\top}$, the optimum is $\mOmega_1=\mU\mV^{\top}$. We complete
$\mOmega_1$ with an orthonormal basis $\mOmega_{\mathrm{pad}}$ of its complement,
$\mOmega=[\mOmega_1\mid\mOmega_{\mathrm{pad}}]$, and the component encoder is the transpose,
$g_{\Omega}(\vz)=\mOmega^{\top}(\vz-\bar{\vz})$. The first $p$ components are aligned with the
annotated concepts and the remaining $D-p$ span everything else.

\para{Unsupervised decomposition (SVD).}
Without annotations we take the right singular vectors of the representations themselves,
$\mZ=\mU\mSigma\mV^{\top}$ and $\mOmega=\mV$, again with
$g_{\Omega}(\vz)=\mOmega^{\top}(\vz-\bar{\vz})$. These are directions of decreasing variance, not
of guaranteed semantics; DiDAE is what makes them inspectable, by rendering a controlled move
along each of them.

\para{Sparse overcomplete decomposition (SAE).}
An orthogonal basis cannot hold more concepts than dimensions. A sparse
autoencoder~\cite{bricken2023towards} lifts this restriction with an overcomplete dictionary,
$D'>D$, and sparse coefficients $\mC\in\mathbb{R}^{N\times D'}$ that reconstruct the
representations under a sparsity constraint, for instance at most $K$ active atoms per sample,
\begin{equation}
    \min_{\mOmega,\mC}\ \|\mZ-\mC\mOmega^{\top}\|_F^2
    \quad\text{s.t.}\quad
    \|\vc_i\|_0\le K\ \ \forall i,
    \label{eq:sae_objective}
\end{equation}
where $\vc_i$ is the coefficient vector of sample $i$. The learned sparse encoder is
$g_{\Omega}$ and the decoder weights are the atoms. Because the atoms are not orthogonal, an
intervention (Eq.~\ref{eq:component_intervention}) changes one coefficient and leaves the others
and the reconstruction residual untouched, which is an atom-wise edit rather than a change of one
independent coordinate. Which SAEs we train, and which public one we reuse, is stated in
Section~\ref{sec:model_architectures}.

\subsection{From Direction to Edit}
\label{sec:component_interventions_live}
\label{subsec:linesearch_factor}

Eq.~\ref{eq:component_intervention} leaves open how far to move. Two fixed choices suggest
themselves: \emph{projection}, $c'_k=0$, which removes the component (or neutralises it when $0$
is its center), and \emph{reflection}, $c'_k=-c_k$, which inverts it when its distribution is
symmetric about $0$. Both are instances of $c'_k=(1-l)\,c_k$ with a line-search factor $l$ of $1$
and $2$, and any fixed $l$ is too weak for some samples and pushes others off the data support.
We instead let the data set the step. On a validation set $\gD_{\mathrm{val}}$ we record the
empirical range of every component,
\begin{equation}
    c_k^{\min}
    =\min_{\vx\in\gD_{\mathrm{val}}}
    \big[g_{\Omega}(\Phi(\vx))\big]_k,
    \qquad
    c_k^{\max}
    =\max_{\vx\in\gD_{\mathrm{val}}}
    \big[g_{\Omega}(\Phi(\vx))\big]_k,
    \label{eq:empirical_component_bounds}
\end{equation}
and an edit moves the coefficient to the bound on the requested side: $c'_k=c_k^{\max}$ to
increase the component and $c'_k=c_k^{\min}$ to decrease it. This is the single edit drawn in
Figure~\ref{fig:overview}c, and Figure~\ref{fig:dae_vs_didae}b shows why it stays axis-parallel
where a global method moves diagonally. This is a trust region rather than
a manifold guarantee: the coefficient stays inside the range the data support, while the complete
edited representation is not constrained further. It removes the global step size altogether,
and the per-sample factor it implies, $l=(c_k-c'_k)/c_k$ for $c_k\neq0$, is what the ablation of
Section~\ref{qualitative_comparison} compares against fixed values.

\para{Classifier-aware edits.}
To explain a downstream classifier $f$ we first distil it into a linear probe on the frozen
representation,
\begin{equation}
    P(\vz)=\vw^{\top}\vz+b,
    \qquad
    P(\Phi(\vx_i))\approx f(\vx_i),
    \label{eq:distilled_probe}
\end{equation}
fitted once, offline, with its fidelity to $f$ evaluated separately; if $f$ already is a linear
probe on $\Phi$, nothing needs to be distilled. The probe then decides only \emph{which} bound to
move to. An edit $\delta_k\vv_k$ changes the probe score by $\delta_k\,\vw^{\top}\vv_k$, so the
score moves towards the boundary when
$\operatorname{sign}(\delta_k)=-\operatorname{sign}\big(P(\vz)\,\vw^{\top}\vv_k\big)$, and the
target coefficient is the empirical bound on that side. The step that would land exactly on the
boundary,
\begin{equation}
    \delta_k^{\star}
    =-\frac{P(\vz)}{\vw^{\top}\vv_k},
    \label{eq:boundary_projection}
\end{equation}
does not set the step size. It enters only as a test: the edit crosses the boundary in
representation space exactly when $|c'_k-c_k|\ge|\delta_k^{\star}|$, which is the
\emph{latent flip} the ranking mode of Section~\ref{sec:ranking-didae} screens its candidates with.
A direction nearly orthogonal to $\vw$ makes the denominator vanish, so we floor its magnitude
at a small constant $\tau$ to keep the test defined; such a direction is bounded by the trust
region rather than excluded.

\para{Pair edits.} For an overcomplete dictionary a single coefficient often cannot carry a
sample across the boundary within its trust region. A \emph{pair edit} therefore moves two
atoms at once, decreasing $c_a$ to $c_a^{\min}$ and increasing $c_b$ to $c_b^{\max}$ in one
step, $\vz'_{\mathrm{FM}}=\vz_{\mathrm{FM}}+(c_a^{\min}-c_a)\vv_a+(c_b^{\max}-c_b)\vv_b$; the
pair $(a,b)$ is then treated as one direction everywhere below (written $a\!\rightarrow\!b$ in
Figure~\ref{fig:ranking-results}).

\subsection{Inversion and Counterfactual Decoding}
\label{sec:inversion}

The decoder of Section~\ref{sec:conditional_score_field} generates from noise; to edit a
\emph{given} image we need the latent code that generates it. An inversion $\operatorname{Inv}$
maps an image and its representation to such a code, $\vx_T=\operatorname{Inv}(\vx,\vz_{\mathrm{FM}})$,
and the decoder $\operatorname{Dec}$ maps a code and a representation back to an image. The
pipeline needs two properties of this pair. First, decoding under unchanged conditioning returns
the input, $\operatorname{Dec}(\vx_T,\vz_{\mathrm{FM}})\approx\vx$. Second, decoding under the
edited conditioning, $\tilde{\vx}=\operatorname{Dec}(\vx_T,\vz'_{\mathrm{FM}})$, transports the
edit to pixels while the code carries what $\vz_{\mathrm{FM}}$ does not determine, such as pose,
layout and texture. This division of labour is what makes a single-vector edit produce a
counterfactual of \emph{this} image rather than a fresh sample with the edited attribute; it is
empirical, not an orthogonality guarantee.

Two inversions meet the requirements. \emph{DDIM inversion}~\cite{song2020denoising} integrates
the deterministic probability-flow ODE of the diffusion model backwards from the image to a single
noise state $\vx_T$, and forwards again under the edited conditioning. It is deterministic but
lossy: the discretization error appears as smoothing and loss of high-frequency detail, which is
tolerable at low resolution. \emph{Edit-friendly DDPM inversion}~\cite{huberman2024edit} draws
the noisy states independently and stores, for every step, the noise map the reverse process
needs to reproduce the next state; the code is then $\vx_T$ together with these maps, and decoding
replays them under the edited conditioning. The reconstruction property holds exactly by
construction, which is why it is the default behind every quantitative result.
Both schemes, and the transfer of the second to a flow-matching schedule, are stated in
Appendices~\appref{app:ddim_inversion}, \appref{app:ddpm_inversion} and~\appref{app:rae_inversion}. The rest of the pipeline only sees the interface, written
$\operatorname{Inv}$ and $\operatorname{Dec}$ in the algorithms, so a more accurate inversion can
be substituted without touching anything else; which inversion each dataset uses is listed in
Table~\ref{tab:components}.

\subsection{Disentangled Diffusion Autoencoders and Direction Ranking}
\label{sec:didae_assemled}
\label{sec:ranking-didae}

Algorithm~\ref{alg:unified_reflection} assembles the pieces. The decoder, the dictionary, the
empirical ranges and, when a classifier is to be explained, the distilled probe are in place
beforehand. Per image, DiDAE inverts once and reuses the code for every direction, so a set of
counterfactuals along several components costs one inversion and one decoding per component. No
gradient is ever taken and no per-image optimization runs, which is where the speed-up over
gradient-based explainers comes from.

\para{Ranking a dictionary against a classifier.} Run over a dataset instead of a single
image, the same loop answers a different question: which directions of the dictionary does a
given classifier actually use? Browsing atoms by their highest-activating inputs
(Section~\ref{section:related}, Figure~\ref{fig:teaser}a) answers this by correlation and for no
classifier in particular. DiDAE answers it by intervention, and the lines marked
\emph{ranking} in Algorithm~\ref{alg:unified_reflection} are all this takes.
Given a trained classifier~$f$ and a \emph{pretrained} dictionary in the frozen encoder space, it
returns a short ranked list of the directions that causally drive~$f$ (Figure~\ref{fig:teaser}b). In ranking mode the distilled probe of Eq.~\ref{eq:distilled_probe} is not fitted in the raw
representation space but by \emph{lasso regression along the components of the sparse
dictionary}: with $\hat{\mOmega}$ the dictionary with unit-normalized atoms,
$\va_i=\hat{\mOmega}^{\top}(\Phi(\vx_i)-\bar{\vz})$ the coefficients of image $i$ and
$y_i=f_1(\vx_i)-f_0(\vx_i)$ the signed margin of $f$, we solve
\begin{equation}
    \boldsymbol{\beta},b\leftarrow\argmin_{\boldsymbol{\beta},b}\ \frac{1}{N}\sum_{i=1}^{N}
    \big(\boldsymbol{\beta}^{\top}\va_i+b-y_i\big)^2+\lambda\|\boldsymbol{\beta}\|_1,
    \qquad \vw=\hat{\mOmega}\boldsymbol{\beta},
    \label{eq:lasso_probe}
\end{equation}
and map the sparse coefficients back to the representation. The $\ell_1$ penalty leaves the few
atoms the probe actually reads as candidates, $\mathcal{K}=\{k:\beta_k\neq0\}$ (Step~0);
$\lambda$ is chosen from a sweep of the support size (Table~\appref{tab:hyper_editing}). Each candidate is then
\emph{verified} rather than scored: the latent-flip test of Eq.~\ref{eq:boundary_projection}
decides which edits are worth decoding (Step~2.4), the decoded counterfactual is re-classified
with the \emph{real}~$f$, and a direction earns a \emph{verified ambient flip} only when that
decision changes \emph{and} a re-encoding confirms the intended component moved, that is, the
sign of $[g_{\Omega}(\Phi(\tilde{\vx}))]_k-c_k$ equals the sign of the requested displacement
(Step~4). Directions are ordered by their verified flips.

The output is therefore classifier-specific, ranked and interventional: no SAE is trained, atoms
that top-activate without mattering are discarded, the evidence is a before/after pair rather than
a gallery, and the same directions, once a teacher has labeled them, feed the CFKD iteration of
Section~\ref{sec:didae_cfkd} that repairs the classifier (Figure~\ref{fig:teaser}c). Every
image whose edit passes the latent-flip screen is decoded; there is no decode budget.

\begin{algorithm*}[t]
\caption{Disentangled Diffusion Autoencoder (DiDAE): counterfactuals for one image and, optionally, a ranking of the dictionary}
\label{alg:unified_reflection}
\begin{algorithmic}
\STATE {\bfseries Input:} image $\vx$ (a dataset $\{\vx_i\}_{i=1}^{N}$ when ranking); frozen encoder $\Phi$; decoder and its inversion
$(\operatorname{Dec},\operatorname{Inv})$; dictionary $\mOmega=[\vv_1,\ldots,\vv_{D'}]$ with
component encoder $g_{\Omega}$; empirical ranges $[c_k^{\min},c_k^{\max}]$
(Eq.~\ref{eq:empirical_component_bounds}); components of interest $\mathcal{K}$; denominator floor $\tau$
\STATE {\bfseries Optional input:} distilled probe $P(\vz)=\vw^{\top}\vz+b$ of a classifier $f$
(Eq.~\ref{eq:distilled_probe}); when ranking, the classifier $f$ itself and the $\ell_1$ penalty
$\lambda$, and the probe is fitted in Step~0 instead of given
\STATE {\bfseries Output:} counterfactuals $\{\tilde{\vx}_k\}_{k\in\mathcal{K}}$; when ranking, $\mathcal{K}$ ordered by verified flips $n_k^{\mathrm{ver}}$ with the flipped before/after pairs
\IF{ranking}
    \STATE $\va_i\leftarrow\hat{\mOmega}^{\top}(\Phi(\vx_i)-\bar{\vz})$, \; $y_i\leftarrow f_1(\vx_i)-f_0(\vx_i)$ for all $i$ \hfill \COMMENT{\textbf{Step 0: lasso regression} of $f$ along the dictionary components, Eq.~\ref{eq:lasso_probe}}
    \STATE $\boldsymbol{\beta},b\leftarrow\argmin_{\boldsymbol{\beta},b}\ \frac{1}{N}\sum_i(\boldsymbol{\beta}^{\top}\va_i+b-y_i)^2+\lambda\|\boldsymbol{\beta}\|_1$; \quad $\vw\leftarrow\hat{\mOmega}\boldsymbol{\beta}$; \quad $\mathcal{K}\leftarrow\{k:\beta_k\neq0\}$; \quad $n_k^{\mathrm{ver}}\leftarrow0$ \hfill \COMMENT{the probe in the dictionary basis; its support are the candidate atoms}
\ENDIF
\FOR{each image $\vx$ (every $\vx_i$ when ranking)}
\STATE $\vz_{\mathrm{FM}} \leftarrow \Phi(\vx)$; \quad $\vx_T \leftarrow \operatorname{Inv}(\vx,\vz_{\mathrm{FM}})$ \hfill \COMMENT{\textbf{Step 1: encode once}, representation and latent code}
\STATE $\vc \leftarrow g_{\Omega}(\vz_{\mathrm{FM}})$ \hfill \COMMENT{\textbf{Step 2.1: coefficients} in the dictionary}
\FOR{each $k\in\mathcal{K}$}
    \IF{$P$ is provided}
        \STATE $s_k \leftarrow -\operatorname{sign}\big(P(\vz_{\mathrm{FM}})\, d_k\big)$, with $d_k=\vw^{\top}\vv_k$ and $|d_k|$ floored at $\tau$ \hfill \COMMENT{\textbf{Step 2.2: side} towards the boundary}
    \ELSE
        \STATE $s_k \leftarrow$ requested side, by default $-\operatorname{sign}(c_k)$ \hfill \COMMENT{\textbf{Step 2.2: side} towards the component's center}
    \ENDIF
    \STATE $c'_k \leftarrow c_k^{\max}$ if $s_k>0$, else $c_k^{\min}$ \hfill \COMMENT{\textbf{Step 2.3: trust region}, edit to the empirical bound}
    \IF{ranking \textbf{and} $|c'_k-c_k|<|\delta_k^{\star}|$}
        \STATE skip $k$ for this image \hfill \COMMENT{\textbf{Step 2.4: latent-flip screen}, Eq.~\ref{eq:boundary_projection}: the edit stays on the probe's side}
    \ENDIF
    \STATE $\vz'_{\mathrm{FM}} \leftarrow \vz_{\mathrm{FM}} + (c'_k - c_k)\,\vv_k$ \hfill \COMMENT{Eq.~\ref{eq:component_intervention}}
    \STATE $\tilde{\vx}_k \leftarrow \operatorname{Dec}(\vx_T,\vz'_{\mathrm{FM}})$ \hfill \COMMENT{\textbf{Step 3: decode} under the edited conditioning}
    \IF{ranking \textbf{and} $f(\tilde{\vx}_k)\neq f(\vx)$ \textbf{and} $\operatorname{sign}\big([g_{\Omega}(\Phi(\tilde{\vx}_k))]_k-c_k\big)=\operatorname{sign}(c'_k-c_k)$}
        \STATE $n_k^{\mathrm{ver}}\leftarrow n_k^{\mathrm{ver}}+1$ \hfill \COMMENT{\textbf{Step 4: verified ambient flip}, the real classifier changed and the component moved}
    \ENDIF
\ENDFOR
\ENDFOR
\STATE \textbf{return} $\{\tilde{\vx}_k\}_{k\in\mathcal{K}}$; when ranking, $\mathcal{K}$ sorted by $n_k^{\mathrm{ver}}$ descending, each direction with the pairs $(\vx,\tilde{\vx}_k)$ that flipped
\end{algorithmic}
\end{algorithm*}

\subsection{DiDAE-CFKD}
\label{sec:didae_cfkd}

Counterfactual Knowledge Distillation~\cite{bender2023towards} (CFKD,
Appendix~\appref{appendix:cfkd}, Algorithm~\appref{alg:cfkd}) repairs a classifier by augmentation, and it takes DiDAE
counterfactuals from either route: those generated for hand-picked or all dictionary components
(Section~\ref{subsec:linesearch_factor}) and those generated for the directions the ranking
surfaced (Section~\ref{sec:didae_assemled}, Figure~\ref{fig:teaser}c). Two automatic checks run
first and discard counterfactuals that cannot teach anything: the student must actually have been
flipped, and the image must lie in distribution; the thresholds are given in
Appendix~\appref{sec:downstream_eval}, \emph{Downstream correction}. A teacher then inspects each remaining counterfactual and
decides whether it changed the causal feature (a \emph{true} counterfactual, which is discarded)
or only a spurious one (a \emph{false} counterfactual, which is added to the training set with the
label of its factual), and the classifier is fine-tuned on the augmented set. The human-in-the-loop teacher of~\cite{bender2025mitigating} decides per counterfactual,
which for $Q$ directions, $M$ downstream models and $J$ counterfactuals per direction amounts to
$Q\cdot M\cdot J$ decisions. DiDAE-CFKD exploits that every DiDAE counterfactual comes with the
direction that produced it, and that a direction has a fixed meaning in the frozen encoder space.
It offers two cheaper teachers. With \emph{automatic labeling}, available metadata map each
component to a concept once (for Procrustes the mapping is given by construction) and that label
decides for every counterfactual generated along it. With the \emph{pre-clustered teacher},
counterfactuals are grouped by direction, the teacher sees one collage of before/after pairs per
group and labels the \emph{direction}, not the image, as true, false or out of distribution. The
label transfers to every counterfactual along the direction and to every downstream model that
reads the same encoder, so $Q$ decisions replace $Q\cdot M\cdot J$.

\section{Experimental Setup}
\label{sec:experiments}

Table~\ref{tab:components} is the map of this section: one row per dataset with the confounder
its task carries, the instance filling each of the four slots of Section~\ref{section:methods},
and the appendix sections holding the details. The text below gives only what is needed to read
Section~\ref{section:results}. Appendix~\appref{app:experimental_details} has the datasets, splits
and poisoning (\appref{app:datasets}), the encoders, decoders, inversions, dictionaries and students
(\appref{app:instances}) and the four protocols with their thresholds (\appref{app:protocols}); the
metrics are defined in Appendix~\appref{section:desiderata} and the hyperparameters in
Tables~\appref{tab:hyper_generator} and~\appref{tab:hyper_editing}.\ifpreprint\else\footnote{All appendices are provided as supplementary material.}\fi

\begin{table*}[t]
\centering \footnotesize
\caption{\textbf{The six datasets and the instances of the four DiDAE components on each.} The
first three rows carry the counterfactual-quality and correction benchmark, the last three the
dictionary and ranking experiments (Appendix~\appref{app:protocols}, \emph{Dictionary inspection} and \emph{Ranking}). We train the ResNet-18 encoders of the synthetic datasets, the batch
top-$K$ SAEs and the decoders, fine-tuning PathLDM from its histopathology checkpoint; the other
encoders and the MSAE are pretrained. The last column points
to the appendix sections with the dataset and decoder details; resolutions, step counts and
dictionary sizes are in Table~\appref{tab:hyper_generator}.}
\label{tab:components}
\renewcommand{\arraystretch}{1.15}
\setlength{\tabcolsep}{4pt}
\begin{tabular}{llllllll}
\toprule
Dataset & Task & Confounder & Encoder $\Phi$ & Decoder & Inversion & Dictionary & Appendix \\
\midrule
Square          & square intensity     & background      & ResNet-18     & DiffAE  & DDPM & Procrustes / SVD  & \appref{app:datasets}, \appref{app:diffae} \\
CelebA-Blond    & \texttt{Blond\_Hair} & \texttt{Male}   & CLIP ViT-L/14 & DiffAE  & DDPM & Procrustes        & \appref{app:datasets}, \appref{app:diffae} \\
Camelyon17      & tumor                & hospital        & PLIP          & PathLDM & DDPM & batch top-$K$ SAE        & \appref{app:datasets}, \appref{app:pathldm} \\
\midrule
Sparse Numbers  & \texttt{Num128}      & \texttt{Num713} & ResNet-18     & DiffAE  & DDIM & batch top-$K$ SAE & \appref{app:datasets}, \appref{app:diffae} \\
NICO++          & crocodile / lizard   & context         & CLIP ViT-L/14 & RAE     & DDPM & MSAE              & \appref{app:datasets}, \appref{app:rae} \\
ImageNet        & freight / passenger car & found by ranking & CLIP ViT-L/14 & RAE     & DDPM & MSAE              & \appref{app:datasets}, \appref{app:rae} \\
                & fireboat / lifeboat  & water jet (planted) & CLIP ViT-L/14 & RAE     & DDPM & MSAE              & \appref{app:datasets}, \appref{app:rae} \\
\bottomrule
\end{tabular}
\end{table*}

\subsection{Datasets}
\label{sec:experimental_setup}

We evaluate on six datasets with a known spurious correlation (Table~\ref{tab:components}).
Square, a synthetic benchmark whose square intensity is confounded with the background intensity,
CelebA-Blond~\cite{liu2015deep}, where \texttt{Blond\_Hair} is confounded with \texttt{Male}, and
Camelyon17~\cite{koh2021wilds}, where the tumor label is confounded with the hospital, carry the
counterfactual-quality benchmark and the downstream-correction experiments in the protocol
of~\cite{bender2025mitigating}: the training split is poisoned so that the confounder predicts the
label ($98\%$ of the samples on the first two, $100\%$ on Camelyon17) and the test split of
$N=1000$ samples is balanced over class and confounder, so that accuracy reflects the causal
feature.
We perform the dictionary and ranking experiments on Sparse Numbers, a synthetic benchmark of $1{,}000$ three-digit numbers in superposition
(four per image on average, every number in $0.4\%$ of the samples), the \emph{crocodile} vs.\
\emph{lizard} task of NICO++~\cite{zhang2023nico} with the context as the cue, and the ImageNet
\emph{freight car} vs.\ \emph{passenger car} pair, trained on the natural data, and the \emph{fireboat} vs.\ \emph{lifeboat} pair with the documented water-jet shortcut~\cite{neuhaus2023spurious}, which
we plant by filtering the training split with the water-jet atom \texttt{\#5717}
(Appendix~\appref{app:datasets}), so recovering \texttt{\#5717} checks the ranking, it is not a discovery.

\subsection{Model Instances}
\label{sec:model_architectures}

On the two synthetic datasets the encoder is a ResNet-18 trained on the known
generative factors, so that the linear representation hypothesis holds by construction; the
real-image datasets use the OpenAI CLIP ViT-L/14 image encoder~\cite{radford2021learning} or, for
histopathology, PLIP~\cite{huang2023visual}. The Procrustes dictionaries are fitted on the training
split to the annotated attributes, SVD on the same representations, the batch top-$K$
SAE~\cite{bussmann2024batchtopk} is trained by us on Sparse Numbers and Camelyon, and the $6{,}144$-atom
MSAE~\cite{zaigrajew2025interpreting} over CLIP is taken as is; none of them is fitted to a
student. Every quantitative result uses edit-friendly DDPM inversion (Appendices~\appref{app:ddpm_inversion} and~\appref{app:rae_inversion})
and the empirical trust region of Section~\ref{subsec:linesearch_factor}; DDIM inversion appears
only on Sparse Numbers and in the ablation of Table~\appref{tab:generator_inversion_ablation}. We
correct two kinds of \emph{downstream classifiers}: a ResNet-18 trained from scratch on the poisoned split, the
primary subject of the correction experiments, and a \emph{probe}, a linear head on frozen
foundation-model features, which on ImageNet reads DINOv3~\cite{simeoni2025dinov3} ViT-L/16 rather than the dictionary's
CLIP. The \emph{distilled probe} $P$ of Eq.~\ref{eq:distilled_probe} that DiDAE edits against is
fitted once per student in closed form, and by lasso along the dictionary in ranking mode
(Eq.~\ref{eq:lasso_probe}); Appendix~\appref{app:instances} gives every instance in full.

\subsection{Protocols}
\label{sec:protocols}

Four protocols produce the results of Section~\ref{section:results}, in the order they are
reported there; thresholds, budgets and the treatment of the baselines are in
Appendix~\appref{app:protocols}.

\begin{figure*}[t]
    \centering
    \includegraphics[scale=.25]{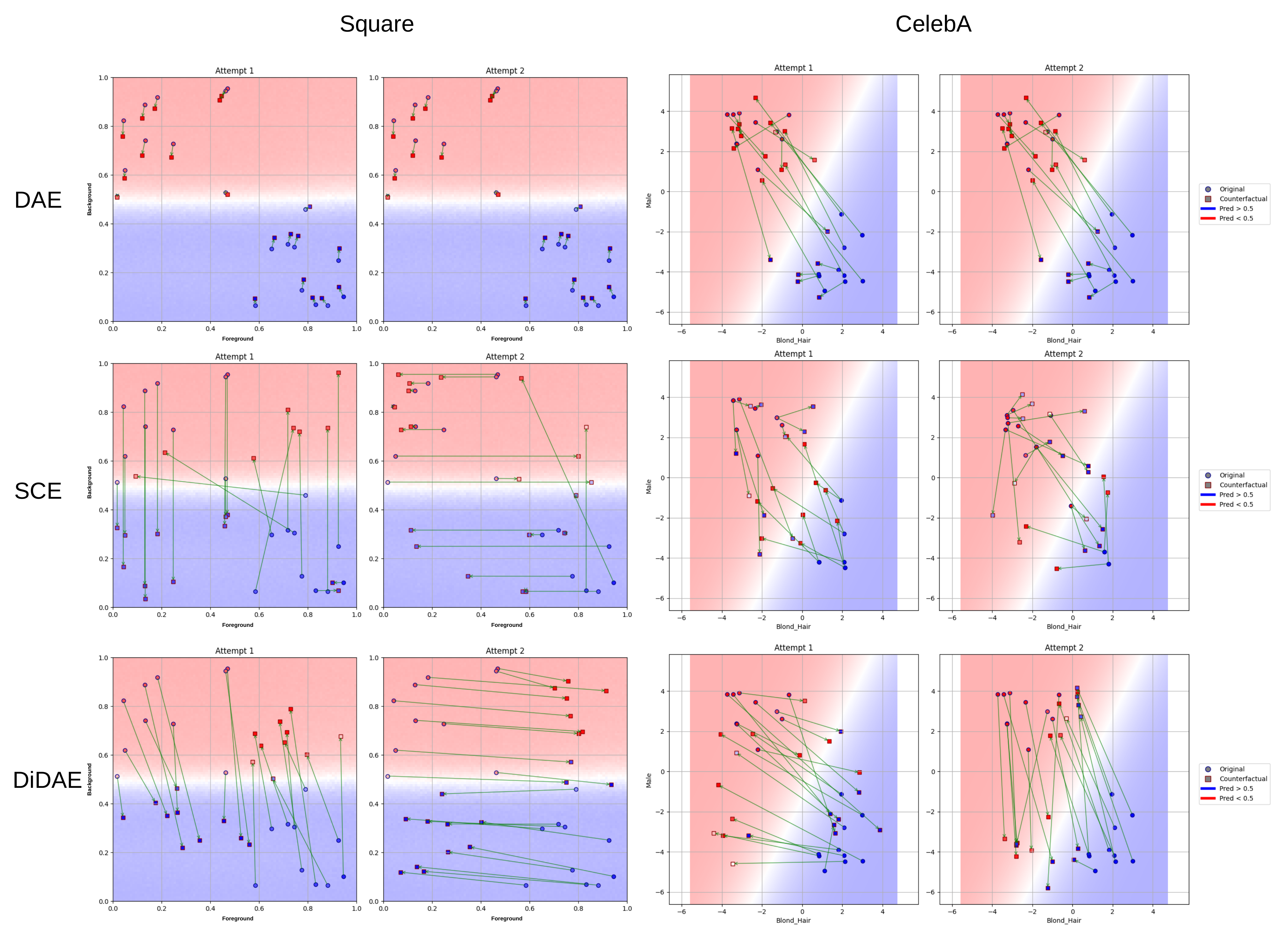}
    \caption{\textbf{DiDAE's two attempts move along different axes; DAE's coincide.} Two
counterfactual attempts per method on Square and CelebA, projected onto the causal (x) and
confounding (y) axes of the oracle encoder defined in Appendix~\appref{section:desiderata}; on Square the decision
boundary is exact, on CelebA it is approximated by the oracle's Male and Blond predictions. DAE
does not cross the boundary on Square and mixes both factors in one direction on CelebA. DiDAE's
trajectories are axis-parallel and roughly orthogonal between attempts, and all counterfactuals of
one attempt share a direction, which is what makes per-cluster teacher feedback possible.}
\label{fig:global_counterfactuals}
\end{figure*}

\para{Counterfactual quality.}
\label{section:methodology}
On Square, CelebA-Blond and Camelyon17 we generate classifier-aware counterfactuals for the
ResNet-18 student (Algorithm~\ref{alg:unified_reflection} with $P$, two attempts per factual) and
compare with DAE~\cite{preechakul2022diffusion}, DiME~\cite{jeanneret2022diffusion},
ACE~\cite{jeanneret2023adversarial}, FastDiME~\cite{weng2024fast} and SCE~\cite{bender2025desiderata}
under the desiderata of Appendix~\appref{section:desiderata} and the downstream Gain of the
correction protocol below. DAE stands in for the gradient-free one-direction family (TIME, GCD,
LeapFactual) and SCE is the strongest gradient-based explainer. We weigh the evidence in order: Gain first, measured on real, held-out, balanced data with group
labels no explainer sees; then the unselected random samples of Appendix Figure~\appref{fig:qualitative_random_ddpm_full} and the projected
trajectories (Figure~\ref{fig:global_counterfactuals}), which show
whether an edit is semantic and single-factor; the desiderata last, as necessary but not
sufficient criteria, since a non-semantic edit that transfers between the models they are computed
with can score high.

\para{Dictionary inspection.}
\label{lab:sae}
Component-defined counterfactuals along every dictionary direction
(Algorithm~\ref{alg:unified_reflection} without $P$) show what each Procrustes and SVD component
encodes on Square and CelebA (Figures~\appref{fig:qualitative_results} and~\appref{fig:svd_didae}) and
what one batch top-$K$ SAE dimension encodes on Sparse Numbers (Figure~\appref{fig:sae_didae});
Procrustes and SVD are also compared by Gain (Table~\ref{tab:didae_cfkd_ablations}).

\para{Downstream correction.}
DiDAE-CFKD (Section~\ref{sec:didae_cfkd}) is applied once to each student with $800$ training and
$200$ validation counterfactuals and one fine-tuning iteration; the teacher is an oracle, an
unpoisoned classifier consulted per counterfactual, unless a table states the pre-clustered
teacher. We report the Gain in average group accuracy on the balanced test split
(Appendix~\appref{app:gain}) and compare with the metadata-based baselines
GroupDRO~\cite{sagawa2020distributionally}, DFR~\cite{kirichenko2023last},
P-ClArC~\cite{anders2022finding} and RR-ClArC~\cite{dreyer2024hope}
(Table~\ref{tab:sota_comparison}), and with projecting the spurious direction out of the frozen
embedding instead of running CFKD (Table~\ref{tab:didae_cfkd_ablations}).

\para{Ranking.}
\label{sec:exp-ranking-didae}
We evaluate the ranking mode of DiDAE (Algorithm~\ref{alg:unified_reflection} with ranking),
called \emph{Ranking DiDAE} in this and the next section, on four classifiers: ResNet-18 students on the two-class Sparse Numbers variant and on NICO++, and
DINOv3 linear probes on the two ImageNet pairs, with dictionaries never fitted to the classifier. We
report the three quantities that define the funnel of Section~\ref{sec:didae_assemled},
\emph{latent flips} (distilled-probe flips in latent space), \emph{ambient flips} (flips of the real
classifier on the decoded image) and \emph{verified flips} (ambient flips whose target component moved
in the requested direction under re-encoding, the ranking key), and the average group accuracy
before and after CFKD on the directions the teacher marks spurious. The teacher is an oracle (an unpoisoned
classifier) on Sparse Numbers and NICO++ and a human, the first author, on ImageNet.

\section{Results}
\label{section:results}

This section answers the questions a practitioner would ask before using DiDAE, each with the table
or figure that settles it\footnote{The results can be reproduced by running
\url{https://github.com/Explainable-AI-Berlin/pytorch_explain_and_adapt_library/blob/441d4d7b323c1437e618dc2c69459051f4315ea9/reproduction_scripts/reproduce_didae_results.sh}.}:
are its counterfactuals good and fast enough (Section~\ref{quantitative_desiderata}), do they change
one concept at a time (Section~\ref{qualitative_comparison}), which dictionary should one use
(Section~\ref{latent_space_exploration}), do they repair a Clever Hans model
(Section~\ref{downstream_model_correction}), and does DiDAE find the shortcut on its own
(Section~\ref{sec:results-ranking-didae})?

\begin{table*}[t!]
\caption{\textbf{Counterfactual quality, speed and downstream Gain.} Edit-friendly DDPM
inversion; desiderata of Appendix~\appref{section:desiderata} in percent, CF/s in counterfactuals per
second on one A100, Gain on the balanced test split. \textbf{Bold} is the best value in a column
within a dataset block, \underline{underline} the second best. Entries are mean~$\pm$~population
std over the $n$ seeds in the third column; $n=1$ entries are a single seed. Gain is the column we
read first (Section~\ref{quantitative_desiderata}).}
\label{tab:quantitative_results_main}
\centering \footnotesize
\renewcommand{\arraystretch}{1.1}
\setlength{\tabcolsep}{4pt}
\begin{tabular}{llrcccccccc}
\toprule
 & & & \multicolumn{6}{c}{\cellcolor{CadetBlue!20}\textbf{desiderata}}\\
 & & & \multicolumn{2}{c|}{\cellcolor{CadetBlue!20}{sufficiency}} & \multicolumn{1}{c|}{\cellcolor{CadetBlue!20}{understandability}} & \multicolumn{2}{c|}{\cellcolor{CadetBlue!20}{fidelity}} & \multicolumn{1}{c|}{\cellcolor{CadetBlue!20}{efficiency}}\\
Dataset & Method & $n$ & \cellcolor{CadetBlue!20}(NAFR) & \multicolumn{1}{c|}{\cellcolor{CadetBlue!20}(Diversity)} & \multicolumn{1}{c|}{\cellcolor{CadetBlue!20}(Sparsity)} & \cellcolor{CadetBlue!20}(NA) & \multicolumn{1}{c|}{\cellcolor{CadetBlue!20}(Unbiasedness)} & \multicolumn{1}{c|}{\cellcolor{CadetBlue!20}(CF/s)} & Gain \\
\multirow{6}{*}{Square}                            & DAE           & 4 & 0.0$\pm$0.0 & 0.0$\pm$0.0 & 0.0$\pm$0.0 & 0.0$\pm$0.0 & 0.0$\pm$0.0 & $\sim$ \textbf{57.1$\pm$2.8} & 0.0$\pm$0.0 \\
                                                   & DiME          & 4 & 8.0$\pm$6.7 & 0.7$\pm$0.6 & 78.6$\pm$3.4 & 32.7$\pm$6.1 & 31.6$\pm$31.7 & $\sim$ 0.03$\pm$0.00 & 69.1$\pm$3.5 \\
                                                   & ACE           & 4 & 1.2$\pm$0.4 & 10.0$\pm$16.9 & 78.3$\pm$16.7 & 58.4$\pm$8.4 & 0.0$\pm$0.0 & $\sim$ 0.03$\pm$0.01 & 38.8$\pm$38.8 \\
                                                   & FastDiME      & 4 & 6.9$\pm$5.2 & 0.9$\pm$1.6 & \textbf{87.7$\pm$6.9} & 41.1$\pm$21.3 & 55.3$\pm$43.6 & $\sim$ 2.4$\pm$0.2 & 66.3$\pm$7.4 \\
                                                   & SCE           & 4 & \textbf{92.1$\pm$5.0} & \textbf{73.3$\pm$42.4} & \underline{84.0$\pm$5.9} & \textbf{93.4$\pm$5.2} & \underline{81.5$\pm$6.2} & $\sim$ 0.02$\pm$0.00 & \textbf{90.6$\pm$1.0} \\
\rowcolor{gray!10}\cellcolor{white}                & DiDAE (ours)  & 4 & \underline{81.9$\pm$6.4} & \underline{66.1$\pm$38.3} & 75.5$\pm$1.0 & \underline{84.3$\pm$6.1} & \textbf{96.5$\pm$2.8} & $\sim$ \underline{47.8$\pm$11.1} & \underline{77.0$\pm$2.2} \\
\midrule
\multirow{6}{*}{CelebA-Blond}                      & DAE           & 4 & 53.6$\pm$2.3 & 0.0$\pm$0.0 & 20.8$\pm$0.3 & \underline{95.4$\pm$2.9} & 24.0$\pm$10.3 & $\sim$ \textbf{15.6$\pm$2.2} & 5.0$\pm$13.8 \\
                                                   & DiME          & 1 & 19.5 & 25.0 & \underline{47.4} & 19.9 & 48.6 & $\sim$ 0.02 & 16.0 \\
                                                   & ACE           & 1 & 25.0 & 20.1 & 46.7 & 32.7 & 55.9 & $\sim$ 0.03 & 7.1 \\
                                                   & FastDiME      & 4 & 8.4$\pm$2.2 & \underline{29.1$\pm$5.3} & 35.3$\pm$10.1 & 28.5$\pm$9.3 & 33.5$\pm$33.5 & $\sim$ 1.5$\pm$0.1 & 13.1$\pm$3.2 \\
                                                   & SCE           & 1 & \textbf{96.5} & 22.8 & 29.5 & \textbf{96.5} & \textbf{77.9} & $\sim$ 0.02 & \underline{23.4} \\
\rowcolor{gray!10}\cellcolor{white}                & DiDAE (ours)  & 4 & \underline{58.9$\pm$1.3} & \textbf{71.8$\pm$3.2} & \textbf{61.1$\pm$0.7} & 82.7$\pm$2.6 & \underline{72.3$\pm$9.3} & $\sim$ \underline{14.3$\pm$0.1} & \textbf{35.5$\pm$2.3} \\
\midrule
\multirow{6}{*}{Camelyon17}                        & DAE           & 4 & 41.8$\pm$3.0 & 0.0$\pm$0.0 & 16.4$\pm$0.5 & \underline{63.3$\pm$4.4} & \underline{88.1$\pm$8.2} & $\sim$ \textbf{1.07$\pm$0.01} & 12.5$\pm$22.2 \\
                                                   & DiME          & 1 & \underline{49.5} & 2.3 & \textbf{47.6} & 53.2 & 43.6 & $\sim$ 0.02 & -3.9 \\
                                                   & ACE           & 1 & 19.0 & 1.1 & 18.0 & 33.6 & \textbf{93.9} & $\sim$ 0.03 & 7.4 \\
                                                   & FastDiME      & 4 & 34.6$\pm$5.3 & \underline{5.1$\pm$1.7} & \underline{38.4$\pm$4.1} & 38.4$\pm$5.2 & 42.2$\pm$6.8 & $\sim$ \underline{0.43$\pm$0.13} & 15.3$\pm$3.1 \\
                                                   & SCE           & 1 & 43.0 & 0.0 & 16.3 & \textbf{78.9} & 76.4 & $\sim$ 0.02 & \underline{22.2} \\
\rowcolor{gray!10}\cellcolor{white}                & DiDAE (ours)  & 4 & \textbf{58.3$\pm$9.9} & \textbf{7.9$\pm$4.9} & 31.5$\pm$2.4 & 61.7$\pm$10.9 & 56.8$\pm$10.2 & $\sim$ 0.35$\pm$0.00 & \textbf{35.0$\pm$10.0} \\
\midrule
\bottomrule
\end{tabular}
\end{table*}

\begin{figure*}[t]
\centering
\includegraphics[width=\textwidth]{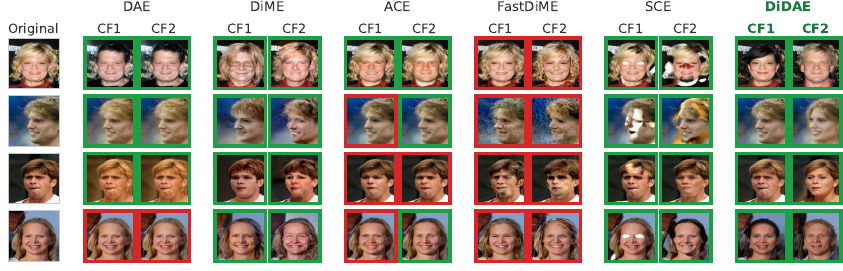}
\caption{Selected qualitative samples on CelebA-Blond under the pipeline of
Table~\ref{tab:quantitative_results_main} (edit-friendly DDPM inversion, seed $0$): four factuals we
chose from a fixed-seed random draw of sixteen of the $200$ shared validation factuals as examples in
which DiDAE's edits flip the classifier. The sixteen unselected factuals, with every attempt of every
method, are in Appendix Figure~\appref{fig:qualitative_random_ddpm_full}, which the evaluation relies
on. Both attempts of every method are shown as stored; a green frame marks an attempt that flipped the
classifier, a red frame one that did not, and for SCE, CF1 is its less aggressive attempt. DAE's two
attempts coincide; DiME and ACE flip through faint, low-contrast changes; FastDiME rarely flips; SCE
flips reliably but often through masks and textures that leave the face manifold, while DiDAE changes
hair color or sex as a photographic edit.}
\label{fig:qualitative_random_ddpm}
\end{figure*}

\begin{table}[t]
\centering \footnotesize
\caption{\textbf{DiDAE-CFKD Gain ablations}, seed $0$, edit-friendly DDPM inversion. The first row
is the reference setting of Table~\ref{tab:quantitative_results_main} (ResNet-18 student, oracle
teacher, Procrustes dictionary, CFKD); every further row changes only the factor named in the first
column. The pre-clustered teacher labels one direction instead of one counterfactual; the probe is
a linear head on the frozen encoder; SVD exists on Square only; projection removes the spurious
direction from the frozen embedding and retrains the probe instead of running CFKD, so its row
compares with the probe row.}
\label{tab:didae_cfkd_ablations}
\renewcommand{\arraystretch}{1.1}
\setlength{\tabcolsep}{3pt}
\begin{tabular}{llcc}
\toprule
Factor & Setting & Square & CelebA-Blond \\
\midrule
Reference  & ResNet-18, oracle, Procrustes & 79.2 & 32.7 \\
\midrule
Teacher    & pre-clustered            & 80.1 & 20.9 \\
Student    & probe                    & 84.7 & 70.6 \\
Dictionary & SVD (ResNet-18)          & 78.2 & --   \\
Correction & projection (probe)       & 53.4 & 22.6 \\
\bottomrule
\end{tabular}
\end{table}

\begin{table}[t]
\centering \footnotesize
\caption{\textbf{Gain against metadata-based baselines.} Square: ResNet-18 students at $98\%$
poisoning; the DiDAE-CFKD row uses the pre-clustered teacher, the weaker of the two teachers in
Table~\ref{tab:didae_cfkd_ablations}. NICO++: the baseline rows are the average-group accuracies
reported in~\cite{bender2025mitigating} for \emph{crocodile} vs.\ \emph{lizard}, converted to Gain; the DiDAE-CFKD entry is the Ranking DiDAE repair of
Figure~\ref{fig:ranking-results}, which corrects a ResNet-18 student like the baseline rows, with
the dictionary living in the frozen CLIP space decoded through the representation autoencoder
rather than in the student's own features.}
\label{tab:sota_comparison}
\renewcommand{\arraystretch}{1.1}
\begin{tabular}{lcc}
\toprule
Method & Square & NICO++ \\
\midrule
GroupDRO & 20.8 &  5.4 \\
DFR      &  2.0 &  0.0 \\
P-ClArC  & 56.2 & -5.0 \\
RR-ClArC & 56.2 & 10.7 \\
\rowcolor{gray!10}\textbf{DiDAE-CFKD} (ours) & \textbf{80.1} & \textbf{16.3} \\
\bottomrule
\end{tabular}
\end{table}

\begin{figure*}[t]
    \centering
    \includegraphics[width=\textwidth]{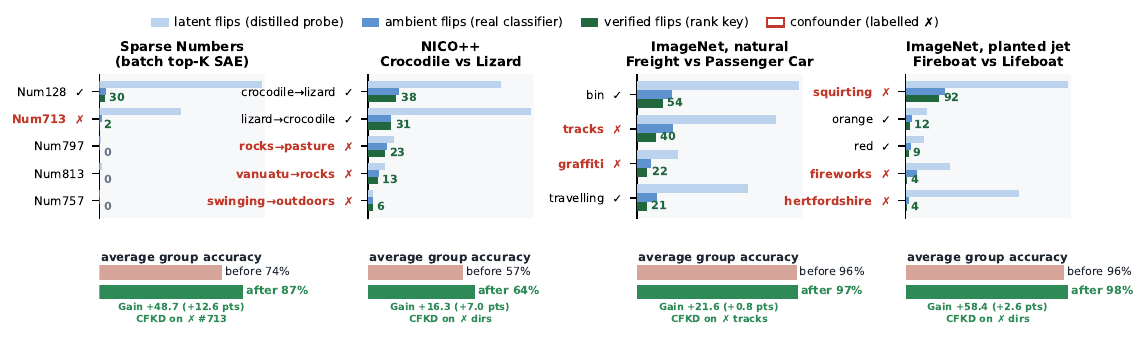}
    \caption{\textbf{Ranking DiDAE surfaces the shortcut, and the funnel narrows at every step.} Top
five directions per classifier (four on freight car, the only ones with verified flips), ordered by verified ambient flips, against a dictionary that already
exists in the frozen encoder space. The bars are the funnel of Section~\ref{sec:exp-ranking-didae}
(linear, per panel), \emph{latent}, \emph{ambient} and \emph{verified} flips (the ranking key, printed
beside the bar); \ding{55} marks directions the teacher calls spurious, \checkmark causal ones. The
lower row gives average group accuracy before and after CFKD on the spurious directions with the
normalized Gain in percent; freight car is the setting of Figure~\ref{fig:teaser}. NICO++ and ImageNet
decode with edit-friendly DDPM inversion at classifier-free guidance $2$, Sparse Numbers with
DDIM.}
\label{fig:ranking-results}
\end{figure*}

\begin{figure*}[t]
    \centering
    \includegraphics[width=\textwidth]{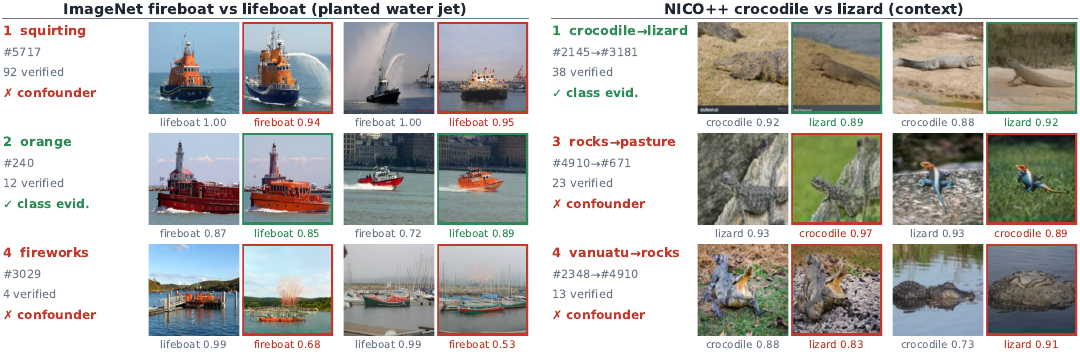}
    \caption{\textbf{Counterfactuals along the top-ranked directions show what the classifier reads.}
Two before/after pairs for three of the directions of Figure~\ref{fig:ranking-results}, chosen by eye from the saved
verified flips, for the planted fireboat probe (left) and the NICO++ student (right), with the
classifier's prediction under each image; the frame gives the teacher's verdict (green class evidence,
red confounder). The fireworks flips are borderline, as its four verified flips suggest.}
\label{fig:ranking-qualitative}
\end{figure*}

\subsection{Desiderata and Efficiency}
\label{quantitative_desiderata}

Table~\ref{tab:quantitative_results_main} scores all methods on the desiderata of Appendix~\appref{section:desiderata}.
Runs of the entire pipeline, from classifier training to the corrected model, are repeated over seeds only where an explainer is cheap enough: DiME, ACE and SCE cost days of GPU time per seed on the real-image and medical datasets, while on Square they carry their four original seeds.
Square DAE flips no sample on any seed, so its desiderata are zero by construction.

\para{Is DiDAE fast and reliable enough to use interactively?} Yes. It produces $47.8$
counterfactuals per second on Square against $0.02$--$0.03$ for the gradient-based explainers
(Table~\ref{tab:quantitative_results_main}), because a counterfactual costs one closed-form latent
edit, one inversion and one decoder rollout rather than hundreds of backward passes. It is also reliable,
with $81.9$/$58.9$/$58.3\%$ NAFR at $84.3$/$82.7$/$61.7\%$ NA, where ACE and FastDiME rarely flip
a sample in a way that transfers ($1.2$--$6.9\%$ NAFR on Square) and DAE, a jointly trained encoder
with a fixed step size and no trust region, fails almost entirely on Square. On Square its weakest column is
sparsity, because an edit moves a coordinate of the representation rather than a region of the image;
on CelebA-Blond, where the factors are global, it is the sparsest method.
NAFR and sparsity also pull against each other when the student reads both features about equally
strongly: a sparse edit changes one, the intact other still votes for the original class and the
flip becomes a coin toss (Appendix Figure~\appref{fig:qualitative_random_ddpm_full}), while an edit that changes
both flips reliably but is not sparse.

\para{Does SCE, the strongest gradient-based explainer, do better?} On some columns, at $17\times$ (Camelyon17)
to about $2{,}400\times$ (Square) the cost. Square, whose two factors are spatially separable, is the ideal
setting for a mask-based sparsifier, and there SCE leads NAFR, diversity, NA and Gain, FastDiME sparsity and
DiDAE unbiasedness. On
CelebA-Blond it flips more often and more transferably (NAFR $96.5$ vs $58.9$, NA $96.5$ vs $82.7$)
while DiDAE leads on diversity, sparsity and Gain; on Camelyon17 SCE keeps NA ($78.9$ vs $61.7$) and
unbiasedness ($76.4$ vs $56.8$) and DiDAE the rest.

\para{So which column decides?} Gain, the only metric whose final measurement is accuracy on real,
held-out, balanced data: it asks whether the counterfactuals, used as CFKD augmentation, make the
student more accurate. The six desiderata are blind to a non-semantic edit that transfers between
the student, the distilled classifier and the oracle they are computed with, as a universal
adversarial pattern does (Appendix~\appref{section:desiderata}, \emph{Reading the columns}); Gain
is immune, because a non-semantic edit teaches the student nothing. On Gain DiDAE
wins clearly on CelebA-Blond ($35.5\pm2.3$ against $23.4$ for SCE), leads on Camelyon17 within its
seed variance ($35.0\pm10.0$ against a single SCE seed at $22.2$) and trails SCE only on Square
($77.0$ against $90.6$). The random samples agree: the masks and textures SCE paints on faces
(Appendix Figure~\appref{fig:qualitative_random_ddpm_full} and Figure~\ref{fig:qualitative_random_ddpm}) are the kind of transferable non-semantic edit that earns
its high NAFR and NA.

\subsection{Disentanglement}
\label{qualitative_comparison}
\label{sec:qualitative_artifacts}

\para{Does a counterfactual change one concept at a time?} DiDAE's do; a global method's cannot.
Global methods move orthogonally to the decision boundary (Figure~\ref{fig:dae_vs_didae}a), and
when the classifier reads a Clever Hans cue, that normal is entangled across the correlated
attributes, so one DAE edit changes hair color, gender and makeup at once. DiDAE moves along one
dictionary axis at a time. Projected onto the causal/spurious plane
(Figure~\ref{fig:global_counterfactuals}), DAE never crosses the boundary on Square and runs
diagonally on CelebA, while DiDAE's trajectories are axis-parallel and orthogonal between attempts.
The unfiltered random samples of Appendix Figure~\appref{fig:qualitative_random_ddpm_full} show the same per image, and
Figure~\ref{fig:qualitative_random_ddpm} shows four of them.

\begin{figure*}[t]
\centering
\includegraphics[width=0.82\textwidth]{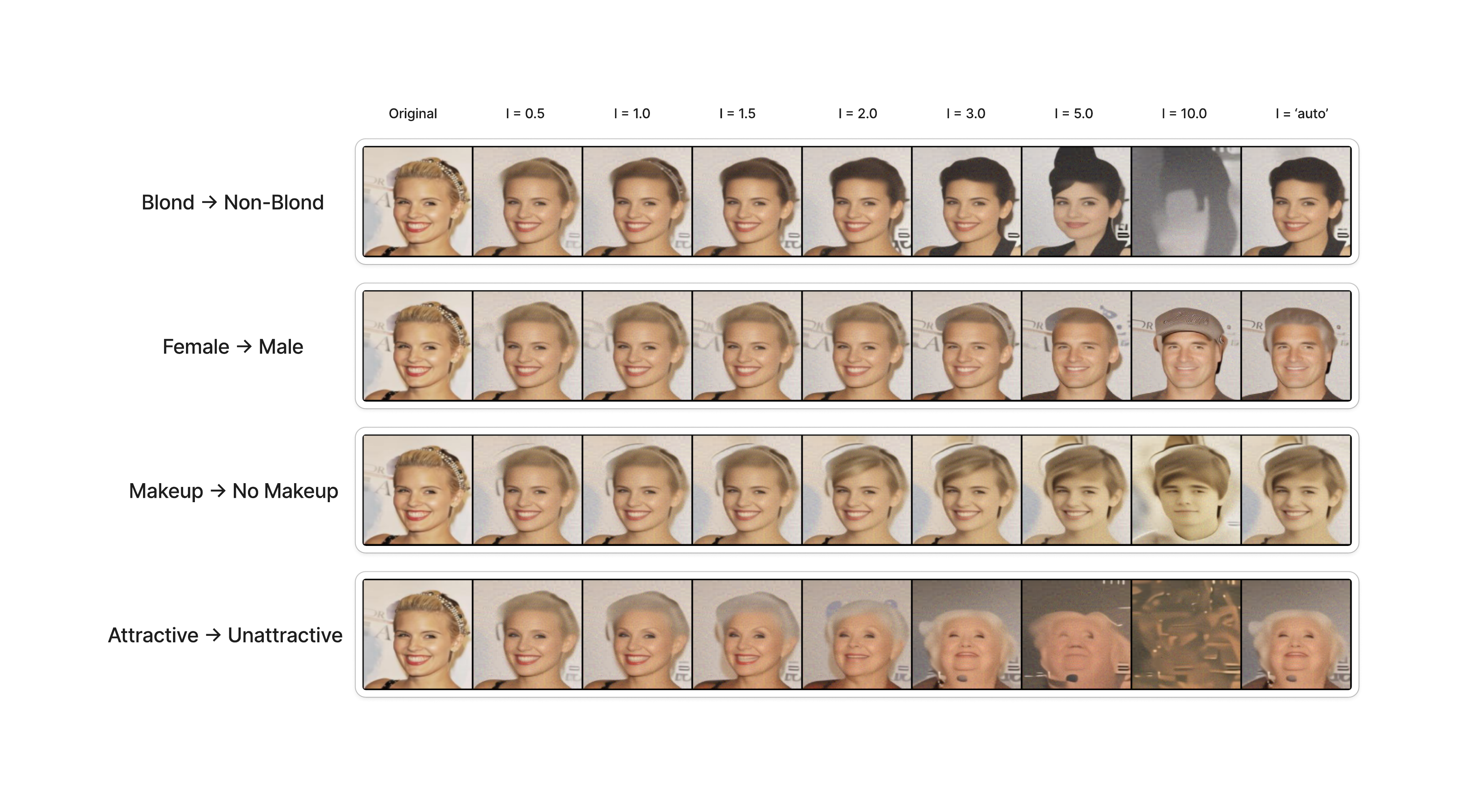}
\caption{Line-search factor ablation on CelebA-Blond: fixed values of $l$ against the empirical trust region (\texttt{auto}) of Section~\ref{subsec:linesearch_factor}. Fixed $l\le1$ barely changes the image and $l\ge5$ leaves the manifold; the trust region does neither. Decoded with DDIM inversion, the weaker of the two inversions of Section~\ref{sec:inversion}.}
\label{fig:didae_linesearch}
\end{figure*}

\para{How far may an edit go?} As far as the adaptive trust region allows: fixed step sizes
$l\le1$ barely change the image and $l\ge5$ leave the manifold, while the trust region does neither
(Figure~\ref{fig:didae_linesearch}), even under the lossy DDIM inversion used there.

\subsection{Dictionaries}
\label{latent_space_exploration}

\para{Do I need concept annotations to build the dictionary?} Not for every student. With
annotations, Procrustes aligns the embedding with the named concepts
(Figure~\appref{fig:qualitative_results}); without them, SVD recovers foreground and background color
on Square but returns the spatial axes rotated (Appendix Figure~\appref{fig:svd_didae}), since
variance fixes the subspace and not the basis within it. On the ResNet-18 student the two are
within a point of Gain (Table~\ref{tab:didae_cfkd_ablations}).

\para{And when concepts are in superposition?} A sparse dictionary separates them: on Sparse
Numbers, deactivating one batch top-$K$ coefficient removes exactly the corresponding number and
leaves its neighbors intact (Figure~\appref{fig:sae_didae}).

\subsection{Downstream Model Correction}
\label{downstream_model_correction}

Table~\ref{tab:didae_cfkd_ablations} varies the pieces behind Gain one at a time around the
reference setting of Table~\ref{tab:quantitative_results_main} at seed $0$: ResNet-18 student,
oracle teacher, Procrustes dictionary, $79.2$ Gain on Square and $32.7$ on CelebA-Blond (the
multi-seed means are $77.0\pm2.2$ and $35.5\pm2.3$).

\para{How does DiDAE-CFKD compare with metadata-based Clever Hans removal?} It beats all four
baselines (Table~\ref{tab:sota_comparison}). Reweighting struggles when the counter-example group is
$2\%$ of the data (GroupDRO: $20.8$ on Square, $5.4$ on NICO++), and metadata-guided unlearning along an
annotated artifact direction (P-/RR-ClArC) works on Square ($56.2$) but barely on NICO++
($-5.0$/$10.7$), where the confounder is the whole background rather than a small artifact. DiDAE-CFKD reaches $80.1$ on Square even with the weaker pre-clustered teacher
and $16.3$ on NICO++ through the Ranking DiDAE repair, without any group labels;
Figure~\appref{fig:didae_cfkd_before_after} shows the decision boundary rotating from the spurious
axis onto the causal one.

\para{Could I just project the spurious direction out instead?} Yes, and that would even fix the
foundation model's representation itself rather than one classifier, but it works far worse: on the
same probe, projection reaches $53.4$ against $84.7$ Gain on Square and $22.6$ against $70.6$ on
CelebA-Blond (Table~\ref{tab:didae_cfkd_ablations}). A foundation model spreads a concept over
redundant directions that removing one dictionary atom does not eliminate, whereas the
counterfactuals teach the probe to ignore the concept wherever it is encoded.

\para{Can the corrected classifier simply be a probe on the foundation model?} Yes, and the Gains
are much larger ($84.7$ on Square, $70.6$ on CelebA-Blond, Table~\ref{tab:didae_cfkd_ablations}),
since the probe reads the representation the edit is made in. It is, however, less flexible, as it
ties the classifier to the edit space, so every other result uses a separate ResNet-18 student, the
more realistic and more conservative estimate.

\para{Does a cheaper teacher suffice?} On Square, yes: the pre-clustered teacher, one label per
direction given by us from the cluster collages instead of one oracle decision per counterfactual
(Section~\ref{sec:didae_cfkd}), costs nothing ($80.1$). On CelebA-Blond it costs about a third of the
Gain ($20.9$), because the directions are less cleanly separated and a cluster-level label admits or
discards a whole direction's counterfactuals at once.

\subsection{Ranking DiDAE}
\label{sec:results-ranking-didae}

\para{Does DiDAE find the shortcut without being told where to look?} Yes, in all four settings of
Figure~\ref{fig:ranking-results}. On \textbf{Sparse Numbers} the causal digit \texttt{Num128} and
the injected confounder \texttt{Num713} take the top two of a ${\sim}10^3$-atom dictionary and
everything else scores zero. On \textbf{NICO++} the two object pairs rank above the three context
pairs. On the natural \textbf{freight car} probe rail tracks and graffiti rank second and third behind the
class evidence ``bin'' (Figure~\ref{fig:teaser}); graffiti is the spurious feature
Neuhaus et al.~\cite{neuhaus2023spurious} report for freight car, the tracks shortcut is new. What an
atom means to the classifier comes from its counterfactuals, not its CLIP-Dissect name: ``tracks''
stands for a view down the rails that often leaves no car at all, and the probe calls such an empty
track a freight car. On the planted
\textbf{fireboat} probe the water jet ranks first, and we mark it, the firework spray and
``hertfordshire'' as confounders. Figure~\ref{fig:ranking-qualitative}
shows what the fireboat and NICO++ directions do: the jet appears or vanishes on an unchanged boat,
and the NICO++ context directions swap rocks for grass behind an unchanged lizard. The ImageNet probes
read DINOv3 features, the dictionary lives in CLIP space: the two need not share a backbone.

\para{Why verify flips instead of counting them in latent space?} Because it changes the ranking:
on fireboat ``hertfordshire'' has $309$ latent flips, second most, but $4$ verify, while the water jet
goes $444\rightarrow109\rightarrow92$.

\para{Does fixing the surfaced directions help?} Yes. CFKD on the directions labeled spurious raises
average group accuracy $74\rightarrow87\%$ on Sparse Numbers (Gain $48.7$), $57\rightarrow64\%$ on
NICO++ ($16.3$), $96.4\rightarrow97.2\%$ on freight car ($21.6$) and $95.5\rightarrow98.1\%$ on fireboat
($58.4$), where on ImageNet the average runs over the four class$\times$confounder groups. The natural
freight car probe errs on only $1$ of $28$ freight and $1$ of $17$ passenger cars with tracks, and the
repair removes a fifth of the remaining group error. On the planted fireboat probe
the improvement lands where the shortcut predicts: images without a water jet rise
$93.4\rightarrow98.1\%$, those with one only $98.8\rightarrow99.3\%$.

\section{Conclusion}
\label{section:conclusion}

We introduced DiDAE, which wraps a frozen foundation model in a conditional diffusion decoder and generates counterfactuals by a closed-form edit along a dictionary direction rather than by iterative gradient search.
The dictionary can be supervised (Procrustes), unsupervised (SVD) or overcomplete and sparse (SAE) without changing the edit operator, which lets the same machinery explain a downstream classifier and the representation space of the foundation model itself.
Run in its ranking mode, it turns an unlabeled dictionary into a short list of the directions a classifier causally reads; on a natural ImageNet probe this surfaced a shortcut not reported before, rail tracks, which one CFKD iteration then reduced.

Moving the search into a dictionary basis buys up to three orders of magnitude in speed, which makes counterfactual-based correction practical at the scale CFKD needs.
The counterfactuals repair downstream models at $77.0\pm2.2$ and $35.5\pm2.3$ Gain on Square and CelebA-Blond, against $20.8$ for GroupDRO on Square.
DiDAE is not uniformly best: SCE stays ahead on Square and on some desiderata columns of the real-image datasets.
DiDAE's case rests first on Gain, measured on real, held-out, balanced data, which it wins clearly on CelebA-Blond and leads within seed variance on Camelyon17; then on the random samples, which show disentangled, on-manifold edits; and only last on the desiderata, which are necessary but not sufficient.

The ranking results carry a separate lesson: verification by decoding and re-classifying reorders the directions, most for the largest dictionary, the regime SAE interpretability operates in.

\section{Limitations}
\label{section:limitations}

\para{DiDAE inherits the quality of its dictionary and its generator.} It is a lens, not a
source of semantics: the dictionary supplies the directions, the generator renders them, and DiDAE
composes and verifies. If the atoms do not correspond to human-recognizable concepts, the edits are
reproducible but not interpretable; verification only confirms that the intended
\emph{coefficient} moved. Both failure modes appear in our results: several ImageNet atoms verify under
CLIP-dissect names too noisy to act on, and the SVD components on Square mix the spatial axes
because variance, not semantics, defines them. 

\para{Minimal flips are harder on ImageNet.} With the representation autoencoder on ImageNet, far
fewer latent flips survive rendering than on CelebA or Square, and many of those that do are not
minimal: the tracks edits of Figure~\ref{fig:teaser} often recompose the whole scene rather than add
rails to it. We attribute this to the data: CelebA and Square are essentially
unimodal, so a straight edit in the representation stays on one connected manifold of faces or
squares; a two-class ImageNet task spans many visual modes whose manifolds need not be connected, and
a flip may require the decoder to jump between them. We have not tested this hypothesis.

\section{Future Work}
\label{section:future_work}

First, the mechanism works on embeddings, so it extends to text, graphs or proteins given an invertible conditional decoder.
Second, one could repair the foundation model rather than each classifier on top of it: fine-tuning it on the labeled pairs the ranking of Section~\ref{sec:ranking-didae} supplies would remove a shortcut once, for every downstream classifier.
Third, a blinded human study of unfiltered counterfactuals of every method would close a gap in our evaluation: the desiderata cannot tell a semantic edit from a transferable adversarial one (Section~\ref{quantitative_desiderata}), and Gain is coarse and teacher-dependent.

\subsubsection*{Acknowledgments}
This work was partly funded by the German Ministry for Education and Research (under refs 01IS14013A-E, 01GQ1115, 01GQ0850, 01IS18056A, 01IS18025A, 13GW0744D, and BIFOLD25B) and by  DFG. Furthermore, K.-R.M. was partly supported by the Institute of Information \& Communications Technology Planning \& Evaluation (IITP) grant funded by the Korea government (MSIT) (No. RS-2019-II190079, Artificial Intelligence Graduate School Program, Korea University) and grant funded by the Korea government (MSIT) (No. RS-2024-00457882, AI Research Hub Project). Lastly, Sidney Bender was partially funded by Bosch-Siemens Haushaltsger\"ate. Correspondence to SN, KRM and MM. Special thanks to David Drexlin for his help with making his representation autoencoders work and to Dominik K\"uhne for his help with setting up the PEAL web demo.

Generative AI systems of the Claude (Anthropic), GPT (OpenAI) and Gemini (Google) families, released 2025--2026~\cite{genai2026}, helped draft or revise the text of all sections, the \textsc{Peal} code and the experiment and figure scripts, and AI agents ran experiments under the authors' direction. The authors designed the method and experiments, labeled the human-teacher directions, checked all content and take full responsibility for it.

\bibliography{iclr2026_conference}
\bibliographystyle{IEEEtran}

\ifpreprint
\appendices
\section{Desiderata of Counterfactuals}
\label{section:desiderata}

We evaluate with the desiderata of~\cite{bender2025desiderata}, which carry the explanation desiderata of Swartout and Moore~\cite{Swartout1993} (fidelity, understandability, sufficiency, efficiency and low construction overhead) over to visual counterfactuals. We measure the first four; the last is hard to quantify. Throughout, $\mathcal{M}$ denotes the manifold of natural images of a dataset.

We report no realism or proximity scores---FID, LPIPS, face-verification similarity,
$\ell_p$ distance---although they are common in the VCE
literature~\cite{jeanneret2022diffusion, jeanneret2023adversarial, weng2024fast}. Distributional
scores judge a set of images rather than a single explanation. Proximity rewards the wrong thing: an edit that spreads many
small changes over the image scores well and tells the user nothing, whereas a readable
counterfactual moves one factor clearly. Semantic sparsity
(Section~\ref{section:understandability}), the oracle measures below and the non-adversarial rate,
which discards flips an independently trained classifier does not reproduce, cover what we need.

\para{Reading the columns.} The six desiderata are classifier- and oracle-mediated: NAFR and NA
are flips of the student and of an independently trained distilled classifier, and diversity,
sparsity and unbiasedness are read off the oracle encoder below. They are therefore structurally
blind to an edit that changes the decision without changing the semantics, as long as the edit
transfers: a universal adversarial pattern that fools every classifier trained on the same data
passes NAFR and NA, and if it also moves the oracle it scores on diversity, sparsity and
unbiasedness as well. Only Gain is immune, because a non-semantic edit teaches the student nothing.
Each column is moreover a marginal: NAFR counts flips of any kind, diversity is averaged only over
the factuals whose two attempts both flipped, sparsity only over flipped attempts, so a method can
score well per column without giving most factuals two orthogonal, on-manifold, single-factor
flips. Two figures therefore belong to the evaluation rather than to its illustration.
Figure~\mainref{fig:global_counterfactuals} projects every attempt of every method onto the oracle's
causal and confounding axes and shows at once whether flips happen, whether the two attempts
separate the factors and whether they run parallel to an axis.
Figure~\ref{fig:qualitative_random_ddpm_full} is the unfiltered check: sixteen validation factuals drawn
with a fixed seed, no exclusions, every attempt of every method, reproducible from the repository;
Figure~\mainref{fig:qualitative_random_ddpm} shows four of them, selected as successful flips.

\begin{figure*}[p]
\centering
\includegraphics[width=\textwidth,height=0.80\textheight,keepaspectratio]{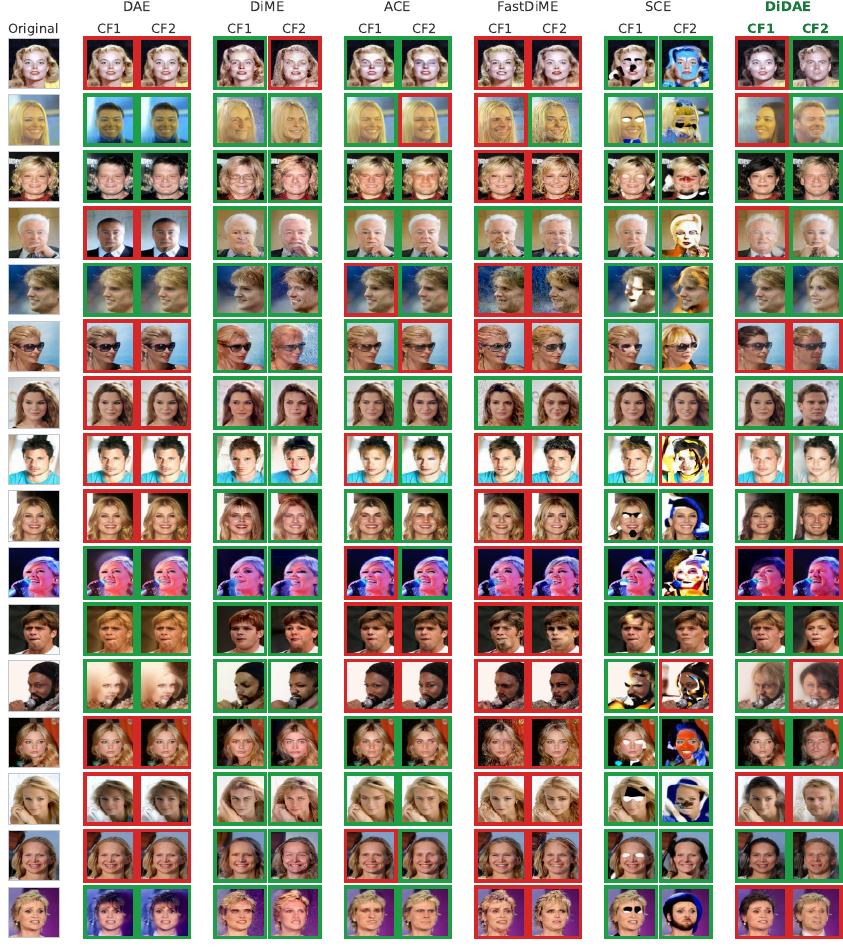}
\caption{Random, unselected qualitative samples on CelebA-Blond under the pipeline of
Table~\mainref{tab:quantitative_results_main} (DiffAE conditioned on OpenAI CLIP ViT-L/14, edit-friendly DDPM
inversion, seed $0$); Figure~\mainref{fig:qualitative_random_ddpm} shows rows 3, 5, 11 and 15, selected as successful flips. The sixteen factuals are drawn uniformly at random with a fixed seed from the
$200$ validation factuals shared by all six runs; both attempts of every method are shown exactly
as generated, without any filtering: a green frame marks an attempt that flipped the
classifier, a red frame one that did not. For SCE, CF1 is its less aggressive attempt. This is the per-sample behavior behind the table: DAE's two attempts coincide, since
it has a single direction per classifier; DiME and ACE flip most samples through faint,
low-contrast changes, which is why their non-adversarial rates in the table are low; FastDiME
rarely flips; SCE flips reliably but frequently through masks and textures that leave the face
manifold, which its high NAFR does not penalize. DiDAE's counterfactuals almost always realize the
intended edit of their semantic dimension (hair color or sex) as a photographic change, so a red
frame on a DiDAE attempt rarely means a failed edit. The poisoned classifier, however, relies on
both dimensions: an edit along one of them leaves the other still pointing to the original class,
the two cues conflict, and whether the decision flips becomes close to a coin toss.}
\label{fig:qualitative_random_ddpm_full}
\end{figure*}

For evaluation only we assume an oracle encoder $\Phi_{\mathrm{orc}}$ mapping an image to two interpretable
dimensions: foreground and background intensity on Square, blondness and maleness on CelebA,
tumor probability and staining intensity on Camelyon17. On Square it is exact and dictionary-free,
reading both intensities off the image using the known position of the square. On CelebA and
Camelyon17 it is CLIP and UNI~\cite{chen2024uni} fine-tuned on group-balanced data to above $90\%$ accuracy on every
group. One caveat follows on CelebA: its oracle is fine-tuned on the same annotations the supervised
Procrustes dictionary is fitted to, so the oracle space is aligned with the axes DiDAE edits
along, which favors DiDAE on diversity, sparsity and unbiasedness. The Camelyon17 dictionary is an
unsupervised SAE and does not share this bias. We found no independent
encoder reliable enough to replace it.

\subsection{Sufficiency}
\label{sec:sufficiency}

A sufficient explanation tells the user enough about the model's behavior to act on it. Classifiers often rely on several intertwined features, so we ask for a \textit{diverse} set of counterfactuals per input $\vx$: if the local decision depends on directions $\vw_1,\dots,\vw_q$, a sufficient explainer returns counterfactuals $\tilde{\vx}_1,\dots,\tilde{\vx}_q$ that each isolate one of them, $|\cos(\vx - \tilde{\vx}_j,\vw_j)| \approx 1$ for all $j$. Following~\cite{bender2025desiderata}, we also tie sufficiency to the non-adversarial flip rate: charting the decision boundary takes flips on many inputs, not one exploitable weakness.

\para{Non-Adversarial Flip Rate (NAFR).}
Bridging both sufficiency and fidelity \cite{bender2025desiderata}, the NAFR tracks the absolute proportion of generated samples that both alter the target model's prediction and induce a genuine semantic shift. We introduce a surrogate classifier $f_{\mathrm{sur}}$, distilled from $f$ but initialized and trained separately, to filter out adversarial artifacts that do not transfer:
\begin{equation}
\begin{aligned}
    \text{Sufficiency:NAFR} ={}& \frac{1}{N} \sum_{i=1}^{N} \mathbb{1}\big( f(\tilde{\vx}_i) = y_{\mathrm{tgt}} \big) \\
    &\quad\cdot\,\mathbb{1}\big( f_{\mathrm{sur}}(\tilde{\vx}_i) = y_{\mathrm{tgt}} \big)
\end{aligned}
\end{equation}
where $y_{\mathrm{tgt}}$ is the target class of the counterfactual, $N$ the number of factuals and $\tilde{\vx}_i$ the attempt for factual $i$ with the highest target confidence of $f$, so NAFR counts factuals, not attempts. This metric effectively isolates functional, visually coherent counterfactuals from those that merely exploit imperceptible gradients.

\para{Diversity.}
To verify \textit{sufficiency}, we assess the explainer's capacity to generate heterogeneous explanations. By generating two distinct counterfactuals for the same factual image, we measure their latent deviation:
\begin{equation}
\text{Sufficiency:Diversity} = 1 - \mathrm{CosSim}(\Delta_1,\Delta_2)
\end{equation}
where $\Delta_1$ and $\Delta_2$ represent the change in the oracle encoder representation of the two respective edits. Higher scores denote a richer and more complete explanatory set.

\subsection{Understandability}
\label{section:understandability}

An explanation must be easy to grasp. A visual counterfactual already lives in the image domain and, ideally, on $\mathcal{M}$; beyond that we ask for \textit{sparsity} of the change $\vx \mapsto \tilde{\vx}$: only a few semantic features should change.

\para{Sparsity.}
To evaluate \textit{understandability}, we measure the conciseness of the semantic change via the Hoyer sparsity of the latent difference vector $\Delta$ of the oracle encoder representations of the factual and the counterfactual:
\begin{equation}
\text{Understandability:Sparsity} = \frac{\sqrt{d_{\mathrm{orc}}} - \frac{|\Delta|_1}{|\Delta|_2}}{\sqrt{d_{\mathrm{orc}}} - 1}
\end{equation}
where $d_{\mathrm{orc}}$ is the dimension of the oracle representation. This formulation yields $1$ for a perfectly sparse edit and $0$ for a completely dense transformation. The final score is averaged across all valid counterfactuals.

\subsection{Fidelity}
\label{section:fidelity}

An explanation must reflect how the model actually decides; a misleading one does harm~\cite{Swartout1993}. For a factual $\vx$ and a classifier $f$ whose local decision boundary has the dominant normal $\vw$, a counterfactual $\tilde{\vx}$ is faithful if it meets three conditions:
\begin{enumerate}[label=(\roman*)]
\item \textbf{On the manifold:} the counterfactual is a realistic image, $\tilde{\vx} \in \mathcal{M}$.
\item \textbf{Neighborhood stability:} The immediate vicinity of the generated sample, defined as $\{\boldsymbol{\xi} \in \mathcal{M}:\|\boldsymbol{\xi} - \tilde{\vx}\|<\varrho\}$ for a small radius $\varrho$, also receives the counterfactual class.
\item \textbf{Directional alignment:} the change is aligned with the local normal, $|\cos(\vx-\tilde{\vx},\vw)| \approx 1$.
\end{enumerate}
These conditions cannot be checked directly in high dimensions with an unknown $\mathcal{M}$, so we use the proxy metrics of~\cite{bender2025desiderata}:

\para{Non-Adversarial Rate (NA).}
To measure \textit{fidelity}, we must ensure counterfactuals occupy robust semantic basins rather than fragile adversarial pockets. We validate the generated images against $f_{\mathrm{sur}}$ from Appendix~\ref{sec:sufficiency}. This measures the transferability of the counterfactual flip, confirming it relies on robust structural features rather than weight-specific noise:
\begin{equation}
\text{Fidelity:NA} = \frac{N_{\text{flipped\_true}}}{N_{\text{flipped}}}
\end{equation}
where $N_{\text{flipped\_true}}$ represents the count of samples that also successfully flip the prediction of the retrained surrogate model. Higher NA values indicate greater semantic stability.

\para{Unbiasedness.}
As a secondary \textit{fidelity} measure, we evaluate whether the explainer prioritizes the primary causal factors (fulfilling the directional alignment criterion). Let $y$ be the model's prediction and $m_1, m_2$ represent indicator variables for two features in the benchmark. We define the dominant feature as $m^\star = \arg\max_m \mathbb{E}[y = m]$ (evaluated where $m_1 \neq m_2$), and the non-dominant feature as $m^-$. Using the oracle encoder $\Phi_{\mathrm{orc}}$, the edit trajectory is $\Delta = \Phi_{\mathrm{orc}}(\vx)-\Phi_{\mathrm{orc}}(\tilde{\vx})$. To equally penalize under- and over-representation of the target concept, we compute the unbiasedness score using a bounded symmetric log-ratio:
\begin{equation}
\begin{aligned}
r &= \tfrac{1}{\nu}\,\mathbb{E}\!\left[\,|\Delta|_{m^\star} \geq |\Delta|_{m^-}\right], \\[2pt]
\text{Fidelity:Unbiasedness} &= \exp\!\left(-\left|\ln r\right|\right).
\end{aligned}
\end{equation}
where $\nu = \mathbb{E}[y=m^\star]$ acts as a normalization factor. This metric strictly penalizes deviations from the ideal alignment ratio.

\subsection{Efficiency}

As a metric for efficiency, we take the speed at which the counterfactuals are calculated:

\para{Counterfactuals per second.}
We call it Efficiency:CF/s in our tables.
On one NVIDIA A100 (80\,GB) we take the largest batch size that fits, generate one batch of counterfactuals, and divide the batch size by the wall-clock time.
This is the per-counterfactual cost only. The one-time construction overhead is comparable across methods: every explainer in Table~\mainref{tab:quantitative_results_main} needs one generative model trained on the dataset, and the diffusion autoencoder DiDAE trains is of the same size as the unconditional diffusion model that DiME, ACE, FastDiME and SCE run their gradients through. Fitting the dictionary and distilling the probe take seconds on top.

\subsection{Gain}
\label{app:gain}
To not only quantify abstract properties of our explainer, but to also have a measure of how actionable these explanations are in the end we add the downstream improvement metric Gain to our metrics.
It comes with the advantage that a positive Gain shows in one number that the explanation is useful at least in the context of a specific downstream task while the other metrics can often be gamed if you are only optimizing for one of them (e.g. using a prototype of the counterfactual class achieves perfect NAFR, only flipping a single simple sample very sparsely and the others not at all achieves a very good sparsity score etc). However, it focuses on a narrow task, and sometimes creating a few reasonable samples for CFKD might be enough for big improvements for simple datasets and e.g. diversity might not even be necessary then.

Consistent with the protocol in \cite{bender2025mitigating}, we use the Average Group Accuracy (AGA) to account for performance disparities across subgroups. The dataset is divided into disjoint groups $\mathcal{G} = \mathcal{Y} \times \mathcal{A}$ defined by the combination of the class label $y$ and the spurious attribute $a$. The AGA is calculated as the unweighted mean of the accuracy on each subgroup:
\begin{equation}
    \text{AGA}(f) = \frac{1}{|\mathcal{G}|} \sum_{G \in \mathcal{G}} \text{Accuracy}_G(f)
\end{equation}
This metric ensures that the model's performance is evaluated equally on minority groups (where spurious correlations fail) and majority groups.

To quantify the effectiveness of our correction strategy, we report the Gain, defined as the percentage of the performance gap closed between the baseline model and the optimal performance (100\%). Unlike a simple accuracy difference, this normalized metric accounts for the varying difficulty of baselines:
\begin{equation}
    \text{Gain} = \frac{\text{AGA}(f_{\text{corrected}}) - \text{AGA}(f_{\text{baseline}})}{1 - \text{AGA}(f_{\text{baseline}})} \times 100
\end{equation}
where $f_{\text{baseline}}$ is the original model trained on poisoned data and $f_{\text{corrected}}$ is the student model after DiDAE-CFKD.

\section{Counterfactual Knowledge Distillation (CFKD)}
\label{appendix:cfkd}

To mitigate the reliance on spurious correlations, we employ Counterfactual Knowledge Distillation (CFKD)~\cite{bender2023towards} with DiDAE as the explainer. CFKD is a data augmentation strategy that distills true causal mechanisms from a teacher into a student classifier $f$ by exposing it to semantically manipulated counterfactuals. 

As detailed in Algorithm~\ref{alg:cfkd}, the CFKD process assumes four primary components: (i) a trained student classifier $f$, (ii) a visual counterfactual explainer (in our case, DiDAE), (iii) a training dataset $\gD$, and (iv) a teacher $\mathcal{T}$ (which can be a human-in-the-loop, an oracle, or our scalable pre-clustered approach). During each iteration, the VCE generates a counterfactual $\tilde{\vx}$ targeted at a specific class $y_{\mathrm{tgt}}$. The teacher then evaluates whether the generation process successfully altered the underlying causal feature (a ``True'' counterfactual) or merely altered non-causal/spurious artifacts (a ``False'' counterfactual). If the causal feature remains unchanged (False), the generated image $\tilde{\vx}$ is injected back into the training dataset with its original factual label $y$, thereby teaching the model to ignore the spurious transformations. Otherwise it is discarded. The classifier $f$ is subsequently fine-tuned on this augmented dataset. Because DiDAE produces highly interpretable, disentangled image-counterfactual pairs at scale, this feedback loop can efficiently correct the student model's decision boundaries without the traditional computational bottlenecks.

\begin{algorithm}[H]
\caption{Counterfactual Knowledge Distillation (CFKD)}
\label{alg:cfkd}
\begin{algorithmic}[1]
\STATE {\bfseries Input:} trained student classifier $f$, training dataset $\gD$, visual counterfactual explainer (DiDAE), teacher $\mathcal{T}$
\STATE {\bfseries Output:} Fine-tuned classifier $f'$
\STATE $\gD_{\mathrm{aug}} \leftarrow \emptyset$
\FOR{each $(\vx, y) \in \mathcal{D}$}
    \STATE Select target label $y_{\mathrm{tgt}} \neq y$
    \STATE Generate counterfactual image $\tilde{\vx}$ using DiDAE based on $\vx$ and $y_{\mathrm{tgt}}$
    \STATE $u \leftarrow \mathcal{T}(\vx, \tilde{\vx})$ \hfill \COMMENT{teacher decides whether $\tilde{\vx}$ is a \textsc{true} or \textsc{false} counterfactual}
    \IF{$u$ is \textsc{false}}
        \STATE $\gD_{\mathrm{aug}} \leftarrow \gD_{\mathrm{aug}} \cup \{(\tilde{\vx}, y)\}$ \hfill \COMMENT{Retain original label}
    \ENDIF
\ENDFOR
\STATE $f' \leftarrow$ fine-tune $f$ on $\gD \cup \gD_{\mathrm{aug}}$
\STATE \textbf{return} $f'$
\end{algorithmic}
\end{algorithm}

\section{Experimental Details}
\label{app:experimental_details}

This appendix expands Section~\mainref{sec:experiments}: the datasets and their splits, the instances
filling the four slots of Section~\mainref{section:methods} together with the students we correct, and
the four protocols behind Section~\mainref{section:results}.

\subsection{Datasets}
\label{app:datasets}

\para{Square.} A synthetic benchmark in which the task is to identify the intensity level of a
small square in the foreground, placed at varying x- and y-positions. The spurious correlation is
injected through the intensity of the background.

\para{CelebA-Blond.} A subset of CelebA~\cite{liu2015deep} with the task of classifying the
attribute \texttt{Blond\_Hair}. The confounder is the attribute \texttt{Male}, which is highly
correlated with the non-blond class in the poisoned training set, so a student learns to read
gender features instead of hair color.

\para{Camelyon17.} A histopathology dataset~\cite{koh2021wilds} with the task of classifying
whether a tissue patch contains a tumor. The confounder is the hospital the patch was collected
in, which is highly correlated with the tumor label in the poisoned training set, so a student
learns hospital-specific staining instead of tissue characteristics.

\para{Sparse Numbers.} A synthetic benchmark of $1{,}000$ concepts in superposition, built to
test sparse dictionaries under controlled assumptions. Each concept is a three-digit number from
$000$ to $999$; a sample shows between one and eight of them in eight pre-allocated slots, each
slot filled with probability ${\approx}0.5$, so a sample carries four numbers on average. Every
number occurs $50$ times in the $12{,}500$ samples, in $0.4\%$ of them. Its two-class variant,
\texttt{Num128} confounded with the distractor \texttt{Num713}, is the synthetic control of the
ranking experiment, where both ground-truth directions are known by construction.

\para{NICO++.} The \emph{crocodile} vs.\ \emph{lizard} task of~\cite{zhang2023nico}, with the
natural context of the animal as the cue.

\para{ImageNet.} The \emph{freight car} vs.\ \emph{passenger car} pair (ImageNet classes $565$ and
$705$, $1{,}300$ images each) is used as it is, and the probe is trained on the natural split. The \emph{fireboat} vs.\ \emph{lifeboat} pair (ImageNet classes $554$ and
$625$, $1{,}300$ images each, split into training, validation and test), whose documented
shortcut is the water jet~\cite{neuhaus2023spurious}. A probe trained on the raw pair reaches
$99.6$--$100\%$ accuracy on the held-out images and leaves nothing to repair, so we make the
shortcut load-bearing on purpose. We filter the \emph{training} split only: a fireboat is kept if
the MSAE water-jet atom \texttt{\#5717} is active on the real image and a lifeboat is kept if it
is inactive, which leaves $1{,}406$ training images; the $520$ validation and test images are the
natural split and still contain jet-free fireboats. The selection uses the atom's activation on
the real image, never a counterfactual.

\para{Poisoning and splits.} For Square and CelebA-Blond we enforce a poisoning ratio of $98\%$:
for $98\%$ of the training samples the spurious attribute is perfectly correlated with the class
label, and the remaining $2\%$ serve as counter-examples. Camelyon17 uses the harder setting of
$100\%$ poisoning, which CFKD is known to handle~\cite{bender2025mitigating}: the minority groups are
absent from the training set altogether. These three are evaluated on a held-out test set of
$N=1000$ samples balanced with respect to class and spurious attribute, so that accuracy reflects
the causal feature rather than adherence to the correlation. Sparse Numbers, NICO++ and ImageNet
are evaluated on their own balanced splits.

\subsection{Model Instances}
\label{app:instances}

Table~\mainref{tab:components} lists the instance filling each slot on every dataset, and the
paragraphs below follow the same order; the last paragraph describes the students on the other
side of the pipeline and how they are distilled.

\para{Encoders.} 1)~\textbf{Square, Sparse Numbers (custom):} since these are synthetic datasets
with known generative factors, we train a ResNet-18 to regress the four ground-truth factors
(position x, position y, foreground intensity, background intensity) on Square and to predict the
numbers present on Sparse Numbers, and use its $512$-dimensional penultimate representation as
$\vz_{\mathrm{FM}}$. This ensures that the linear representation hypothesis actually holds on both
datasets. 2)~\textbf{CelebA, ImageNet, NICO++ (CLIP):} the pretrained OpenAI CLIP ViT-L/14
image encoder~\cite{radford2021learning}. 3)~\textbf{Camelyon17 (PLIP):}
PLIP~\cite{huang2023visual}, whose pathology-aware representations let DiDAE decompose tissue
patches into directions associated with tumor morphology and hospital-specific staining.

\para{Decoders.} The decoder is chosen by domain; all three leave $\Phi$ frozen and are
conditioned on $\vz_{\mathrm{FM}}$, which is all Section~\mainref{section:methods} requires of them,
and Appendix~\ref{app:decoders} gives their architectures and training.
1)~\textbf{Diffusion autoencoder (Square, Sparse Numbers, CelebA):} the pixel-space diffusion
autoencoder of~\cite{preechakul2022diffusion}, at $64\times64$ on the synthetic datasets and
$128\times128$ on CelebA, trained by us. 2)~\textbf{PathLDM (Camelyon17):} a latent diffusion
model pretrained on histopathology~\cite{Yellapragada_2024_WACV}, which we condition on the
frozen PLIP embedding; a decoder trained from scratch on $1$k patches does not reach usable
tissue fidelity. 3)~\textbf{Representation autoencoder (ImageNet, NICO++):} a two-stage
RAE~\cite{zheng2025diffusion} whose flow-matching score network operates on the grid of frozen
CLIP patch tokens rather than on pixels, conditioned on the CLIP image embedding, trained by us.
The trained weights of these natural-image decoders are hosted on Hugging Face and downloaded
automatically by \textsc{Peal}.

\para{Inversion.} Every quantitative result, including all of
Table~\mainref{tab:quantitative_results_main}, uses edit-friendly DDPM inversion
(Appendix~\ref{app:inversion}), which on the RAE is run with classifier-free guidance $2$; DDIM
inversion is used only on Sparse Numbers and in the ablation of
Table~\ref{tab:generator_inversion_ablation}. The empirical trust region of
Section~\mainref{subsec:linesearch_factor}, called \texttt{auto} in
Figure~\mainref{fig:didae_linesearch}, is measured on the validation split and used in every
experiment, with the denominator floor $\tau=10^{-6}$.

\para{Dictionaries.} 1)~\textbf{Procrustes (Square, CelebA):} fitted on the
training split to the annotated generative factors on Square and to the annotated attributes on
CelebA (\texttt{Blond\_Hair}, \texttt{Male}).
2)~\textbf{SVD (Square):} fitted on the same training representations. 3)~\textbf{Batch
top-$K$ SAE (Sparse Numbers, Camelyon17):} trained by us~\cite{bussmann2024batchtopk} on the $512$-d
penultimate representation with $K=4$, the number of concepts a sample carries on average. On Camelyon17 it
is trained on the $512$-d PLIP embedding,
with $1{,}024$ atoms and $K=16$.
4)~\textbf{MSAE (ImageNet, NICO++):} the public $6{,}144$-atom Matryoshka
SAE~\cite{zaigrajew2025interpreting} over OpenAI CLIP ViT-L/14, taken as is. None of the
dictionaries is fitted to a student.

\para{Students and probe distillation.} Three words recur for the models on the other side of
the pipeline. The \emph{student} is the downstream classifier $f$ being explained and corrected.
A \emph{probe} is a student that is itself a linear head on frozen foundation-model features. The
\emph{distilled probe} $P$ of Eq.~\mainref{eq:distilled_probe} is the linear surrogate of $f$ in the
encoder space that DiDAE edits against. We correct two kinds of students.
1)~\textbf{ResNet-18:} a standard CNN trained from scratch on the poisoned training split, the
primary subject of the correction experiments. 2)~\textbf{Probe:} a linear head on the frozen
encoder; on ImageNet it reads DINOv3 ViT-L/16 features, so the probed encoder and the
dictionary's encoder are \emph{different} foundation models. The distilled probe is fitted once
per student in closed form, by least squares of the student's signed margin on the encoder
representations of the training split, $\vw^{\top}\Phi(\vx)+b\approx f_1(\vx)-f_0(\vx)$; for a
probe student on the same encoder, $P=f$ and nothing is distilled. In ranking mode the same
margin is instead regressed by lasso on the coefficients of the sparse dictionary
(Eq.~\mainref{eq:lasso_probe}, solved by iterative soft-thresholding), so that the probe's support
names the candidate atoms.
Table~\mainref{tab:didae_cfkd_ablations} calls the two students ``ResNet-18'' and ``probe''.

\subsection{Protocols}
\label{app:protocols}

The four protocols of Section~\mainref{sec:protocols} in full.

\para{Counterfactual quality.}
On Square, CelebA-Blond and Camelyon17 we generate classifier-aware counterfactuals for the
ResNet-18 student (Algorithm~\mainref{alg:unified_reflection} with $P$, over the dictionary
components of Table~\ref{tab:hyper_generator}, two attempts per factual) and compare with DAE~\cite{preechakul2022diffusion},
DiME~\cite{jeanneret2022diffusion}, ACE~\cite{jeanneret2023adversarial},
FastDiME~\cite{weng2024fast} and SCE~\cite{bender2025desiderata} under the desiderata of
Appendix~\ref{section:desiderata}: non-adversarial flip rate, diversity, sparsity,
non-adversarial rate, unbiasedness and counterfactuals per second, plus the downstream Gain of
the correction protocol below. Baselines use their published defaults, and seeds are repeated
wherever an explainer is cheap enough to rerun. DAE is the representative of the global,
gradient-free family (Section~\mainref{section:related}): TIME, GCD and LeapFactual share its
one-direction-per-classifier structure, none of them diversifies its edits and not all of them
sparsify them, so we benchmark the family through DAE rather than each member. SCE is the strongest
gradient-based explainer and was designed for spatially separable factors, which Square has and the
two real-image datasets do not; we keep it on all three so that this difference is visible. Disentanglement is assessed by projecting the
counterfactual trajectories onto the oracle's causal and confounding axes
(Figure~\mainref{fig:global_counterfactuals}) and qualitatively on unselected random samples
(Figure~\ref{fig:qualitative_random_ddpm_full}).

\para{Dictionary inspection.}
Component-defined counterfactuals along every dictionary direction
(Algorithm~\mainref{alg:unified_reflection} without $P$) show what each Procrustes and SVD component
encodes on Square and CelebA (Figures~\ref{fig:qualitative_results} and~\ref{fig:svd_didae}). On
Sparse Numbers we encode random samples, read off the active SAE dimensions, deactivate one of
them, re-encode the counterfactual, and keep the pair only if exactly the chosen dimension was
deactivated (Figure~\ref{fig:sae_didae}). Procrustes and SVD are also compared by the Gain they
yield under the correction protocol (Table~\mainref{tab:didae_cfkd_ablations}).

\para{Downstream correction.}
\label{sec:downstream_eval}
DiDAE-CFKD (Section~\mainref{sec:didae_cfkd}) is applied once to each student with $800$ training
and $200$ validation counterfactuals, one fine-tuning iteration and the hyperparameters of
Appendix~\ref{app:hyperparameters}. Before the teacher sees a counterfactual it is dropped if the
student already misclassified its factual, if the student's confidence in the target class does
not exceed $0.5$ (the student was not flipped), or if an outlier score marks the image as out of
distribution. The teacher is an oracle, an unpoisoned classifier consulted per counterfactual,
except in Table~\mainref{tab:didae_cfkd_ablations}, where the pre-clustered labels were given once
by us from the cluster collages, so the two teacher settings differ only in the granularity of
the feedback. We report the Gain in average group accuracy on the balanced test split
(Appendix~\ref{section:desiderata}) and compare with the metadata-based baselines
GroupDRO~\cite{sagawa2020distributionally}, DFR~\cite{kirichenko2023last},
P-ClArC~\cite{anders2022finding} and RR-ClArC~\cite{dreyer2024hope}
(Table~\mainref{tab:sota_comparison}), and with projecting the spurious components out of the frozen
embedding and retraining the probe (Table~\mainref{tab:didae_cfkd_ablations}).

\para{Ranking.}
We evaluate the ranking mode of DiDAE (Algorithm~\mainref{alg:unified_reflection} with ranking), which we call \emph{Ranking DiDAE}, on four classifiers: ResNet-18 students on the two-class Sparse
Numbers variant and on NICO++, and DINOv3 linear probes on the two ImageNet pairs of
Section~\mainref{sec:experimental_setup}. On the natural freight car pair the question is which
shortcut, if any, the ranking surfaces among the $6{,}144$ atoms; on the fireboat pair with the planted
water jet, whether the planted atom is recovered without being told which atom was used, and whether
the repair fixes the images without a water jet. Every dictionary already exists in its frozen encoder
space and none is fitted to the classifier under study. On NICO++ we rank concept \emph{pairs}
(Section~\mainref{subsec:linesearch_factor}), which frees the edit from the trust region of a single
coefficient; on ImageNet we rank \emph{single} atoms so that each counterfactual changes one
named concept. The teacher is an oracle (an unpoisoned classifier) on Sparse Numbers and NICO++ and a human, the first author, on
ImageNet, who labels each ranked direction from its before/after pairs. We report the
three quantities that define the funnel: \emph{latent flips} (distilled-probe flips in latent
space), \emph{ambient flips} (flips of the real classifier on the decoded image) and \emph{verified flips}
(ambient flips whose target component moved in the requested direction under re-encoding, the
ranking key). Directions labeled spurious are repaired with CFKD
(Section~\mainref{sec:didae_cfkd}) and we measure average group accuracy on the
balanced test split before and after.

\section{Hyperparameters}
\label{app:hyperparameters}

The hyperparameters for training the pixel-space diffusion autoencoder and for DDIM inversion follow~\cite{ha2025diffusion}. The implementation follows it as well with the change, that we injected and froze the semantic encoder. For running CFKD we used the same hyperparameters as in~\cite{bender2025mitigating} which already had configurations for the ResNet-18 runs. For the runs based on foundation models we just swapped out the model and kept everything else the same. We did not run any hyperparameter search for CFKD on top of this, and all explainers were given the same CFKD budget: two counterfactual attempts per factual (the second drawn with the base seed $+1$), $800$ training and $200$ validation counterfactuals, a generation batch size of $200$, and a single CFKD fine-tuning iteration. The explainer-internal settings of the baselines are the defaults published with each method. Tables~\ref{tab:hyper_generator} and~\ref{tab:hyper_editing} list the remaining choices.

\begin{table*}[t]
\caption{Generator and dictionary configuration per dataset. The Square foundation model is a ResNet-18 trained to regress the four ground-truth generative factors; the dictionary is fitted on its $512$-d penultimate representation rather than on the four regressed factors. $^{\dagger}$DDIM in the ablation of Table~\ref{tab:generator_inversion_ablation} only. $^{\ddagger}$Of the $1024$ atoms, DiDAE edits the two with the largest weight in the distilled probe.}
\label{tab:hyper_generator}
\centering \footnotesize
\begin{tabular}{lcccccc}
\toprule
 & Square & Sparse Numbers & CelebA & Camelyon17 & ImageNet / NICO++ \\
\midrule
Encoder $\Phi$          & ResNet-18 & ResNet-18 & CLIP ViT-L/14 & PLIP & CLIP ViT-L/14 \\
$\dim \vz_{\mathrm{FM}}$  & $512$ & $512$ & $768$ & $512$ & $768$ \\
Generator               & DiffAE & DiffAE & DiffAE & PathLDM & RAE (App.~\ref{app:rae}) \\
Generator resolution    & $64$ & $64$ & $128$ & $256$ & $256$ \\
Classifier resolution   & $64$ & $64$ & $128$ & $128$ & $224$ \\
Inversion               & DDPM$^{\dagger}$ & DDIM & DDPM$^{\dagger}$ & DDPM & DDPM \\
Inversion steps         & $20$ & $20$ & $20$ & $50$ & $50$ \\
Dictionary              & Procrustes / SVD & batch top-$K$ SAE & Procrustes & batch top-$K$ SAE & MSAE \\
Components              & $4$ & $1000$ ($K=4$) & $2$ ($4$ in Fig.~\ref{fig:qualitative_results}) & $1024$ ($K=16$), $2$ edited$^{\ddagger}$  & $6144$ \\
\bottomrule
\end{tabular}
\end{table*}

\begin{table*}[t]
\caption{Editing settings, shared across datasets.}
\label{tab:hyper_editing}
\centering \footnotesize
\begin{tabular}{lp{0.55\textwidth}}
\toprule
Line-search factor $l$ & \texttt{auto}: target set to $c_k^{\min}/c_k^{\max}$ (Eq.~\mainref{eq:empirical_component_bounds}); effective $l$ varies per sample \\
Empirical bounds & measured on the validation split, used unscaled \\
Denominator floor $\tau$ on $\vw^{\top}\vv_k$ & $10^{-6}$ \\
Lasso penalty $\lambda$ (Eq.~\mainref{eq:lasso_probe}, ranking only) & $0.05$ on Sparse Numbers, $0.01$ on NICO++ and ImageNet, chosen from a sweep of the probe's support size, normalized by the representation dimension; $2500$ iterations of iterative soft-thresholding \\
Counterfactual attempts per factual & $2$ \\
CFKD fine-tuning iterations & $1$ \\
Train / validation counterfactuals & $800$ / $200$ \\
Generation batch size & $200$ \\
Teacher & oracle (unpoisoned classifier), including the NICO++ ranking; pre-clustered in Table~\mainref{tab:didae_cfkd_ablations}; human (first author) for the ImageNet ranking \\
Seed & $0$; Table~\mainref{tab:quantitative_results_main} varies it \\
\bottomrule
\end{tabular}
\end{table*}

Two consequences of Table~\ref{tab:hyper_generator} are worth making explicit. First, the CelebA Procrustes dictionary used for the quantitative results spans only the two annotated attributes that define the task (\texttt{Blond\_Hair}, \texttt{Male}); the four-component variant of Figure~\ref{fig:qualitative_results} adds makeup and attractiveness for visualization only. Second, the CelebA generator and its student classifier both operate at $128\times128$, so no resampling step sits between them, and the smoothing of DDIM decoding (Section~\mainref{sec:inversion}) is the inversion's doing rather than a resolution artifact.

\section{Decoder Architectures and Training}
\label{app:decoders}

All three decoders are conditional score networks in the sense of
Section~\mainref{sec:conditional_score_field}: each is trained with the encoder $\Phi$ frozen and
receives $\vz_{\mathrm{FM}}$ as conditioning, and they differ only in what they denoise and how
the conditioning enters the network.

\subsection{Diffusion Autoencoder}
\label{app:diffae}

For Square, Sparse Numbers and CelebA the decoder is the pixel-space diffusion autoencoder
of~\cite{preechakul2022diffusion}. The noise-prediction network $\epsilon_\theta$ is a U-Net, and
the timestep $t$ and the frozen conditioning $\vz_{\mathrm{FM}}$ are injected through adaptive
group normalization (AdaGN) layers, which scale and shift the normalized feature maps
$\vh\in\mathbb{R}^{C\times H\times W}$ channel-wise:
\begin{equation}
    \operatorname{AdaGN}(\vh,t,\vz_{\mathrm{FM}})
    =\vz_s\odot\big(\vt_s\odot\operatorname{GroupNorm}(\vh)+\vt_b\big),
    \label{eq:adagn}
\end{equation}
where $\vz_s=\operatorname{Affine}(\vz_{\mathrm{FM}})\in\mathbb{R}^{C}$ and
$(\vt_s,\vt_b)=\operatorname{MLP}(\psi(t))\in\mathbb{R}^{2\times C}$ is the output of a multilayer
perceptron applied to the sinusoidal timestep encoding $\psi(t)$. Because $\vz_{\mathrm{FM}}$ is
frozen, the score network learns only the image variation along the semantics the pretrained
encoder already represents. Training hyperparameters follow~\cite{ha2025diffusion}
(Appendix~\ref{app:hyperparameters}); resolutions are $64\times64$ on the synthetic datasets and
$128\times128$ on CelebA, where the student classifier operates at the same resolution so that no
resampling step sits between them.

\subsection{PathLDM}
\label{app:pathldm}

For Camelyon17 the decoder is PathLDM~\cite{Yellapragada_2024_WACV}, a latent diffusion model
pretrained on histopathology at $256\times256$. Its score network is conditioned on the frozen
PLIP embedding $\vz_{\mathrm{FM}}$ and trained on the Camelyon17 training patches with $\Phi$
frozen; inversion and decoding run in its latent space with $50$ edit-friendly DDPM steps
(Appendix~\ref{app:ddpm_inversion}).

\subsection{Representation Autoencoder}
\label{app:rae}

For ImageNet and NICO++ the conditional decoder of Section~\mainref{sec:conditional_score_field} operates on a \emph{representation} latent rather than on pixels, following the two-stage representation-autoencoder recipe of~\cite{zheng2025diffusion}. Both stages keep the foundation encoder $\Phi$ (OpenAI CLIP ViT-L/14) frozen; only the decoder and the transformer are trained.

\para{Stage 1: latent decoder.}
The stage-1 latent is the grid of frozen CLIP patch tokens, $\mH\in\mathbb{R}^{16\times16\times1024}$ at $256\times256$ input resolution, and a ViT-XL decoder $\Gamma$ trained from scratch maps it back to pixels. Training minimizes
\begin{equation}
\begin{aligned}
\mathcal{L}_{\text{stage-1}}
  =\;& \|\vx-\Gamma(\mH)\|_1
  + \lambda_{\text{p}}\,\mathcal{L}_{\text{LPIPS}}\big(\vx,\Gamma(\mH)\big) \\
  &+ \lambda_{\text{d}}\,\mathcal{L}_{\text{GAN}}\big(\Gamma(\mH)\big),
\end{aligned}
\end{equation}
with $\lambda_{\text{p}}=1.0$ and $\lambda_{\text{d}}=0.75$ (hinge discriminator loss, vanilla generator loss), the discriminator enabled from epoch $8$ and updated from epoch $6$. We train for $16$ epochs at global batch size $512$ with AdamW ($\text{lr}=2\cdot10^{-4}$, $(\beta_1,\beta_2)=(0.9,0.95)$, no weight decay), a cosine schedule with one warmup epoch decaying to $2\cdot10^{-5}$, and EMA decay $0.9978$. Latent noise of scale $0.8$ regularizes stage-1 training and is disabled once the decoder is frozen for stage 2.

\para{Stage 2: flow-matching transformer.}
Stage 2 models the stage-1 latent with a decoupled diffusion transformer ($1.24$B parameters; depth $[28,2]$, hidden size $[1440,2048]$, $[20,16]$ heads, patch size $1$), trained by flow matching with an $\vx$-prediction parameterization and a logit-normal time distribution. The conditioning is the $768$-d CLIP image embedding $\vz_{\mathrm{FM}}=\Phi(\vx)$---exactly the vector the dictionaries of Section~\mainref{sec:disentangled_component_analysis} decompose---injected through additive AdaLN with classifier-free-guidance dropout $0.1$ against a learned null embedding. We train $46$ epochs over a precomputed latent cache (original and horizontally flipped views, ${\approx}118$M samples) at global batch size $1024$ with AdamW ($\text{lr}=2\cdot10^{-4}$, $(\beta_1,\beta_2)=(0.9,0.95)$), linear warmup over $14$ epochs decaying to $2\cdot10^{-5}$ by epoch $29$, gradient clipping $1.0$ and EMA decay $0.9995$. Sampling uses $50$ steps.

Because stage 2 is conditioned on precisely the embedding DiDAE edits, an edited $\vz'_{\mathrm{FM}}$ is decodable by the same procedure as in the pixel-space diffusion autoencoder, and Algorithm~\mainref{alg:unified_reflection} applies unchanged.

\section{Inversion Schemes}
\label{app:inversion}

Section~\mainref{sec:inversion} uses an inversion only through its interface: $\operatorname{Inv}$
maps an image and its representation to a latent code, $\operatorname{Dec}$ maps the code and a
(possibly edited) representation back to an image, and decoding under the unchanged representation
reconstructs the input. This appendix states the two instances we use and the transfer of the
second to a flow-matching schedule. Throughout, $\bar\alpha_t$ is the cumulative signal
coefficient of the noise schedule (Section~\mainref{sec:conditional_score_field}),
$\epsilon_\theta(\vx_t,t,\vz)$ the conditional noise prediction, and
$\hat{\vx}_0(\vx_t,t,\vz)=\big(\vx_t-\sqrt{1-\bar\alpha_t}\,\epsilon_\theta(\vx_t,t,\vz)\big)/\sqrt{\bar\alpha_t}$
the implied estimate of the clean image.

\subsection{DDIM Inversion}
\label{app:ddim_inversion}

The DDIM sampler~\cite{song2020denoising} is the Euler discretization of the probability-flow
ODE of the diffusion model, and can therefore be run in either direction. Inversion integrates it
from $t=0$ to $T$ under the original conditioning $\vz_{\mathrm{FM}}=\Phi(\vx_0)$; for a step
from $t$ to $t+\Delta t$,
\begin{equation}
\begin{aligned}
    \frac{\vx_{t+\Delta t}}{\sqrt{\bar\alpha_{t+\Delta t}}}
    ={}&
    \frac{\vx_t}{\sqrt{\bar\alpha_t}}
    \\
    &+
    \left(
        \sqrt{\frac{1-\bar\alpha_{t+\Delta t}}{\bar\alpha_{t+\Delta t}}}
        -
        \sqrt{\frac{1-\bar\alpha_t}{\bar\alpha_t}}
    \right)
    \epsilon_\theta(\vx_t,t,\vz_{\mathrm{FM}}),
\end{aligned}
    \label{eq:ddim_inversion}
\end{equation}
which maps $\vx_0$ deterministically to the code $\vx_T$. Decoding integrates the same update
from $T$ back to $0$, starting at $\vx_T$ but conditioned on the edited representation:
\begin{equation}
\begin{aligned}
    \frac{\vx_{t-\Delta t}}{\sqrt{\bar\alpha_{t-\Delta t}}}
    ={}&
    \frac{\vx_t}{\sqrt{\bar\alpha_t}}
    \\
    &+
    \left(
        \sqrt{\frac{1-\bar\alpha_{t-\Delta t}}{\bar\alpha_{t-\Delta t}}}
        -
        \sqrt{\frac{1-\bar\alpha_t}{\bar\alpha_t}}
    \right)
    \epsilon_\theta(\vx_t,t,\vz'_{\mathrm{FM}}).
\end{aligned}
    \label{eq:ddim_decoding}
\end{equation}
With $\vz'_{\mathrm{FM}}=\vz_{\mathrm{FM}}$ the two integrations cancel up to discretization
error, which is the source of the smoothing visible in the DDIM-decoded figures. We use $20$
uniformly spaced steps.

\subsection{Edit-Friendly DDPM Inversion}
\label{app:ddpm_inversion}

The edit-friendly stochastic inversion of~\cite{huberman2024edit}, adapted to the semantic
conditioning of the diffusion autoencoder, avoids the discretization error entirely. Instead of
a single trajectory, we draw the noisy states \emph{independently} from the forward marginals,
\begin{equation}
\begin{aligned}
\vx_t &= \sqrt{\bar\alpha_t}\,\vx_0 + \sqrt{1-\bar\alpha_t}\,\vepsilon_t, \\
\vepsilon_t &\sim\mathcal{N}(\mathbf{0},\mI)\ \text{i.i.d.},\quad t=1,\dots,T,
\end{aligned}
\end{equation}
so consecutive states are not linked by the usual Markov chain. We then read off, for every step,
the noise map the reverse process would have needed in order to land on $\vx_{t-1}$ from $\vx_t$
under the \emph{original} conditioning:
\begin{equation}
\vn_t \;=\; \frac{\vx_{t-1}-\vmu_t\big(\vx_t,\hat{\vx}_0(\vx_t,t,\vz_{\mathrm{FM}})\big)}{\sigma_t},
\label{eq:ef_extract}
\end{equation}
where $\vmu_t$ and $\sigma_t^2$ are the mean and variance of the DDPM posterior
$q(\vx_{t-1}\mid\vx_t,\vx_0)$. The pair $(\vx_T,\{\vn_t\}_{t=1}^{T})$ is the code returned by
$\operatorname{Inv}$.

Decoding runs the reverse chain with those stored noise maps but with the \emph{edited}
conditioning:
\begin{equation}
\vx_{t-1} \;=\; \vmu_t\big(\vx_t,\hat{\vx}_0(\vx_t,t,\vz'_{\mathrm{FM}})\big) \;+\; \sigma_t\,\vn_t .
\label{eq:ef_sample}
\end{equation}
With $\vz'_{\mathrm{FM}}=\vz_{\mathrm{FM}}$ this reconstructs $\vx_0$ up to numerical error by
construction, and with an edited representation it yields a counterfactual that retains the
input's ``noise identity''. We use $T=20$ steps with uniform spacing on the diffusion
autoencoder and $50$ on PathLDM.

\subsection{Stochastic Inversion on the Flow-Matching Schedule}
\label{app:rae_inversion}

The stage-2 transformer of Appendix~\ref{app:rae} is trained by flow matching on a linear
interpolation schedule rather than a variance-preserving diffusion, so
Eqs.~\ref{eq:ef_extract}--\ref{eq:ef_sample} do not transfer verbatim. We restate them for the
schedule
\begin{equation}
\vx_t=(1-t)\,\vx_0+t\,\vepsilon,\qquad t\in[0,1],
\end{equation}
where $t=1$ is noise and $t=0$ is data, and the network is read in the velocity parameterization
$\vu=(\vx_t-\vx_0)/t$, from which $\hat{\vx}_0$ and the implied noise estimate $\hat{\vepsilon}$
follow.

Reading this interpolation as a Markov chain
$\vx_t=\kappa\,\vx_{t'}+\sqrt{t^2-\kappa^2t'^2}\,\vepsilon'$ with $\kappa=(1-t)/(1-t')$ for a
step $t\to t'$ ($t'<t$) gives the posterior noise scale
\begin{equation}
\sigma(t,t')\;=\;\eta\,t'\,\sqrt{1-\Big(\tfrac{\kappa\,t'}{t}\Big)^{2}},
\qquad \sigma(t,t')\le t',
\end{equation}
with $\eta$ interpolating between the deterministic sampler ($\eta=0$) and the fully stochastic
one ($\eta=1$, our default). One sampler step is
\begin{equation}
\vx_{t'}=\underbrace{(1-t')\,\hat{\vx}_0+\sqrt{t'^{2}-\sigma^{2}}\;\hat{\vepsilon}}_{\textstyle \vmu(t,t')}\;+\;\sigma\,\vn,
\end{equation}
and inversion extracts $\vn=(\vx_{t'}-\vmu(t,t'))/\sigma$ exactly as in Eq.~\ref{eq:ef_extract},
on a uniform grid of $50$ values of $t$ from $1$ to $0$.

Two schedule-specific knobs matter in practice. \emph{Classifier-free guidance} is applied during
inversion \emph{and} sampling, $\hat{\vu}=\vu_\emptyset+\gamma(\vu_{\vz}-\vu_\emptyset)$ against the
learned null embedding, so that reconstruction is preserved at $\gamma=1$ while $\gamma>1$
amplifies precisely the conditioning difference an edit introduces; the NICO++ and ImageNet runs
use $\gamma=2$. \emph{Inversion strength} $t_{\text{start}}\le1$ controls how much of the image the
inverted state carries: at $t_{\text{start}}=1$ that state is nearly sufficient on its own and an
edit to $\vz_{\mathrm{FM}}$ is largely ignored at decode time, so lower values, or fresh noise maps
as in SDEdit~\cite{meng2022sdedit}, hand more of the content back to the conditioning at the cost of input fidelity.

\section{Additional Results}
\label{app:additional}

\begin{table*}[t!]
\caption{Inversion and generator ablation on seed $0$, holding the dictionary, probe and CFKD
budget fixed. DiffAE rows (our pixel-space diffusion autoencoder, not the DAE baseline) change only the
sampler; the RAE row swaps the pixel-space diffusion
autoencoder for a representation autoencoder trained on CelebA (CLIP ViT-L/14 encoder, DDPM
inversion; the recipe of Appendix~\ref{app:rae} with a $0.41$B stage-2 transformer trained for $24$
epochs instead of $46$, used at epoch $20$). The first column is the student's flip rate on the validation counterfactuals, counted per attempt,
whereas the NAFR of Table~\mainref{tab:quantitative_results_main} counts a factual once if its best attempt
flips both classifiers, so it can exceed this rate; the other columns are the seed-$0$ values that enter
Table~\mainref{tab:quantitative_results_main}, and the DDPM rows are the reference setting of
Table~\mainref{tab:didae_cfkd_ablations}. That the Square DDIM row runs at $47.8$ CF/s, the four-seed
DDPM mean of Table~\mainref{tab:quantitative_results_main}, is a coincidence.}
\label{tab:generator_inversion_ablation}
\centering \footnotesize
\renewcommand{\arraystretch}{1.1}
\begin{tabular}{llccccccc}
\toprule
Dataset & Inversion & Flip rate & Diversity & Sparsity & NA & Unbiasedness & CF/s & Gain \\
\midrule
\multirow{2}{*}{Square} & DiffAE / DDIM & \textbf{55.2} & 79.2 & 71.9 & \textbf{90.3} & 84.6 & 47.8 & \textbf{88.6} \\
                        & DiffAE / DDPM & 50.5 & \textbf{83.0} & \textbf{76.3} & 79.8 & \textbf{92.8} & \textbf{60.3} & 79.2 \\
\midrule
\multirow{3}{*}{CelebA-Blond} & DiffAE / DDIM & 44.5 & 52.5 & 52.5 & 75.0 & \textbf{87.1} & \textbf{17.3} & 21.3 \\
                              & DiffAE / DDPM & \textbf{53.0} & \textbf{75.8} & 61.2 & 84.6 & 84.9 & 14.2 & \textbf{32.7} \\
                              & RAE / DDPM & 48.8 & 70.2 & \textbf{64.0} & \textbf{84.8} & \textbf{87.1} & 5.0 & 18.3 \\
\bottomrule
\end{tabular}
\end{table*}

\para{Inversion ablation.} Table~\ref{tab:generator_inversion_ablation} shows that the better
inversion is dataset-dependent. On CelebA edit-friendly DDPM wins five of seven columns against DDIM, which keeps unbiasedness and speed, most
clearly on diversity ($+23.3$) and Gain ($+11.4$); on Square DDIM retains a $9.4$-point Gain
advantage even though DDPM is ahead on diversity, sparsity and unbiasedness. Square's two factors
are spatially separable and its decoder operates at $64\times64$, where DDIM's discretization
error costs little, whereas the CelebA edits are global and benefit from exact reconstruction.
The RAE row is a generator comparison rather than a pure inversion ablation, since encoder and
decoder change together; the two Procrustes dictionaries are fitted to the same attribute matrix,
so the edited directions correspond. The RAE is competitive on the desiderata (best sparsity, NA
and unbiasedness) but decodes at a third of the speed and reaches about half the Gain.

\para{Additional figures.} Figures~\ref{fig:qualitative_results} and~\ref{fig:sae_didae} show the component-defined counterfactuals along the Procrustes directions and one deactivated SAE dimension discussed in Section~\mainref{latent_space_exploration}.
\begin{figure*}[t]
    \centering
    \begin{minipage}[b]{0.48\textwidth}
        \centering
        \includegraphics[width=\textwidth]{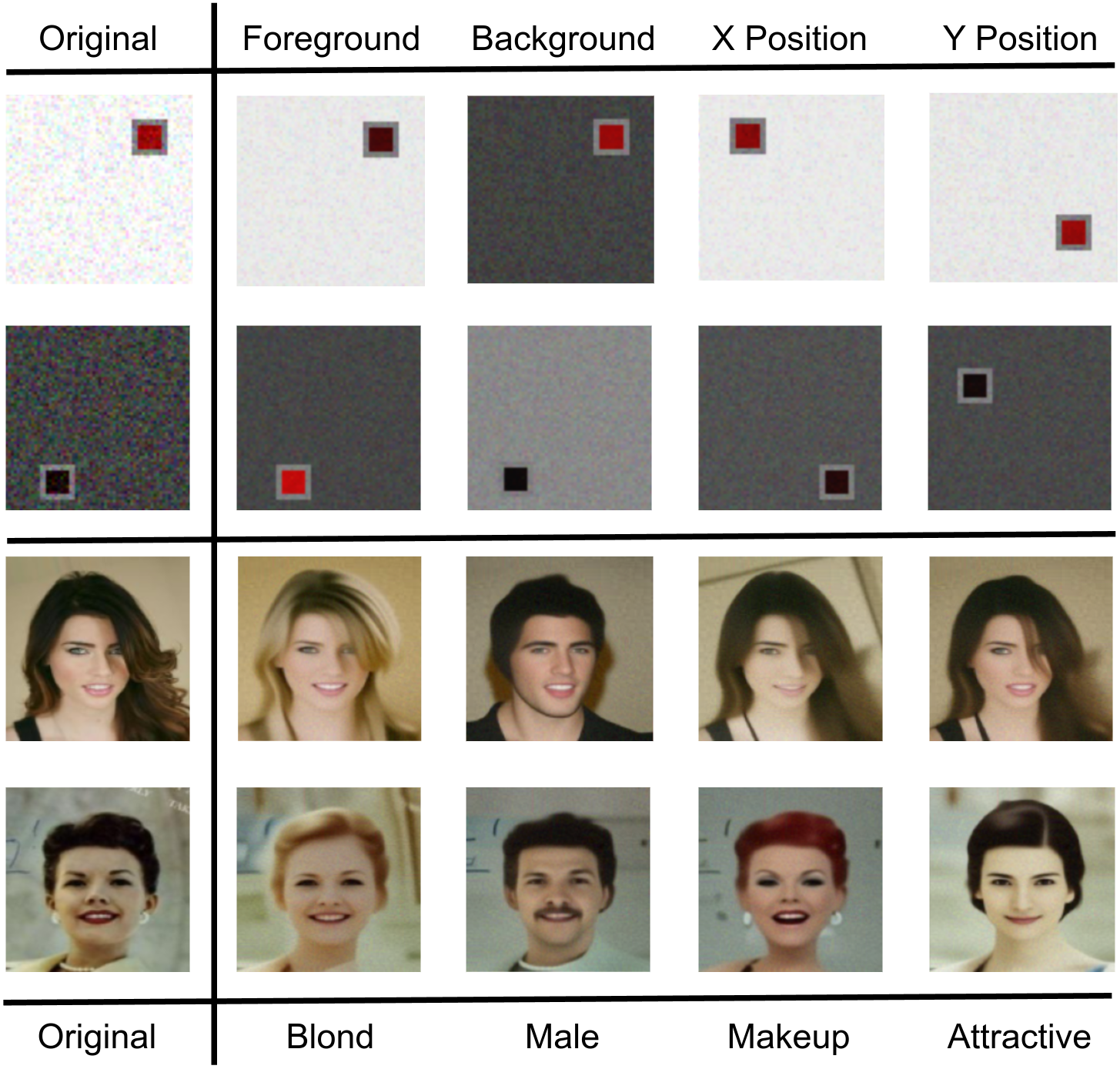}
        \caption{Component-defined counterfactuals along four Procrustes directions. On Square the components are the four generative factors (foreground, background, x position, y position); on CelebA they are \texttt{Blond\_Hair}, \texttt{Male}, \texttt{Heavy\_Makeup} and \texttt{Attractive}, the last three correlated with each other.}
        \label{fig:qualitative_results}
    \end{minipage}
    \hfill 
    \begin{minipage}[b]{0.48\textwidth}
        \centering
        \includegraphics[width=\textwidth]{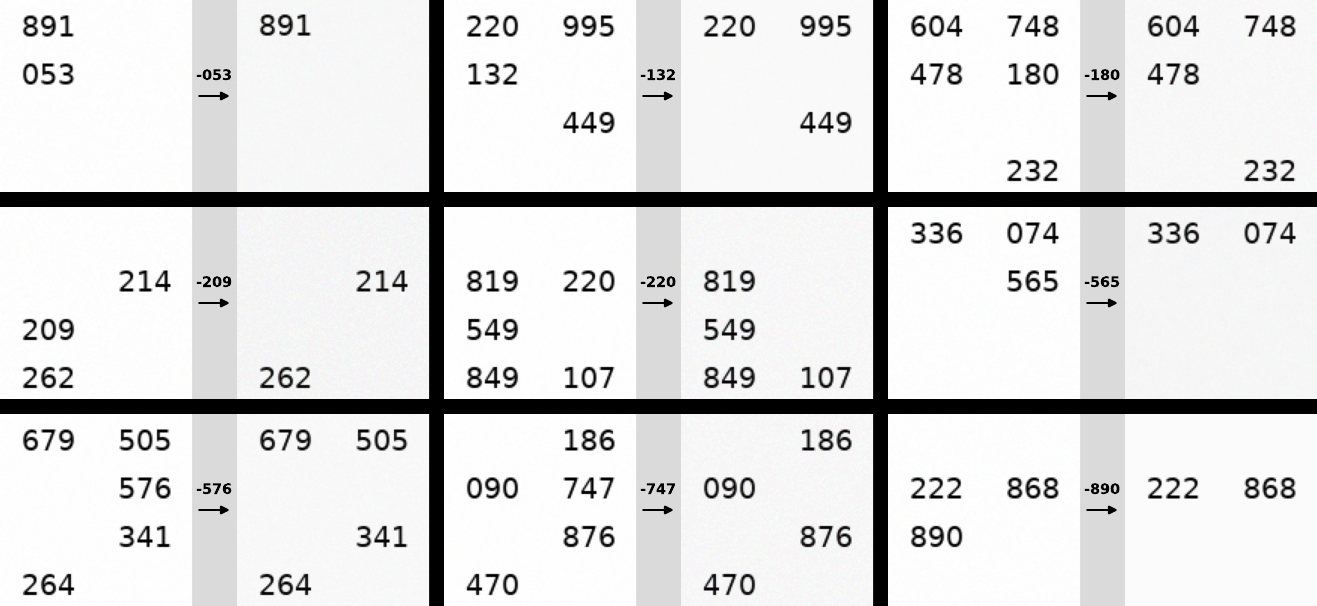}
        \caption{Deactivating one batch top-$K$ SAE dimension with DiDAE on Sparse Numbers. In each column the factual is on the left and the counterfactual on the right; the deactivated dimension is the number that disappears while its neighbors stay.}
        \label{fig:sae_didae}
    \end{minipage}
\end{figure*}
Figure~\ref{fig:svd_didae} shows the counterfactuals along the unsupervised SVD dictionary on
Square discussed in Section~\mainref{latent_space_exploration}.
Figure~\ref{fig:didae_cfkd_before_after} shows how DiDAE-CFKD rotates the decision boundary from
the confounding onto the causal axis; one CFKD iteration was used throughout, although further
iterations are known to help~\cite{bender2023towards}.
\begin{figure}[t]
\centering
\includegraphics[width=\columnwidth]{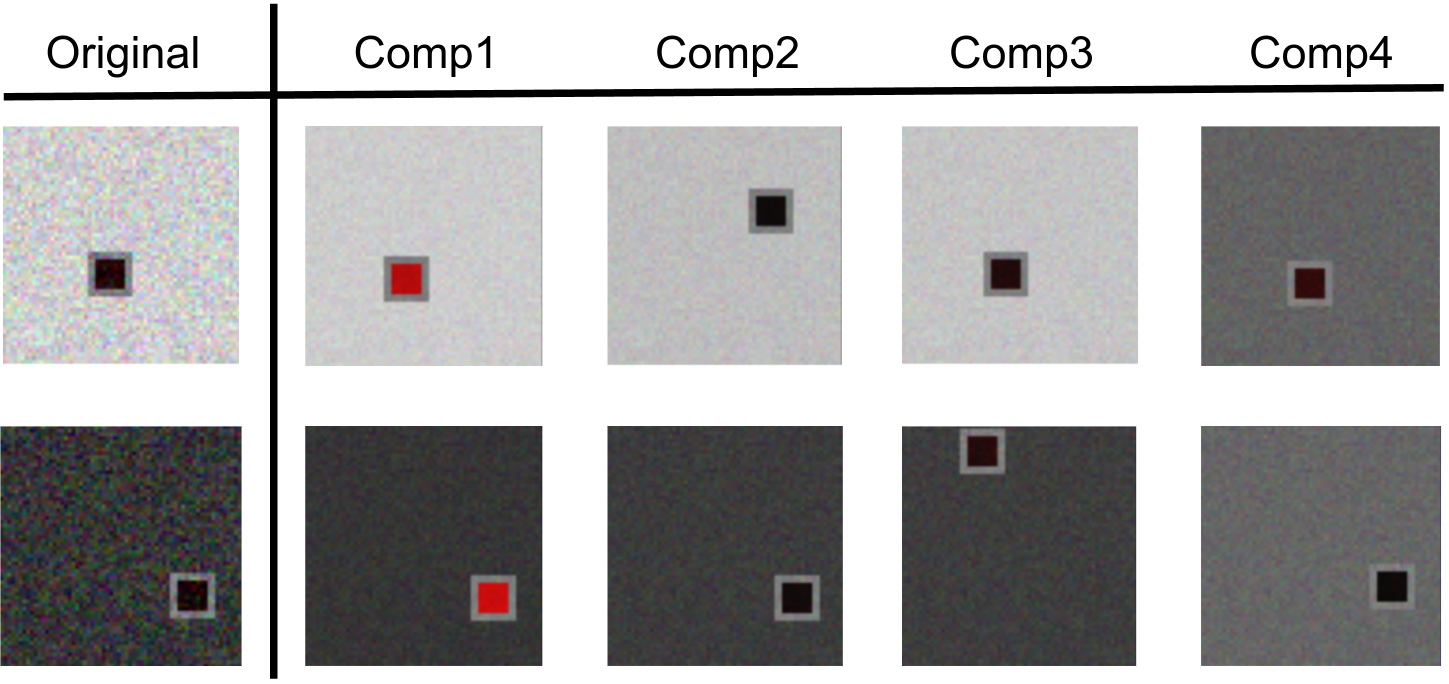}
\caption{Counterfactuals along the first four SVD directions of the Square foundation-model space (Section~\mainref{latent_space_exploration}). Comp1 is the foreground color and Comp4 the background color; Comp2 and Comp3 mix the two spatial axes, which variance alone cannot separate.}
\label{fig:svd_didae}
\end{figure}

\begin{figure*}[t]
\centering
\includegraphics[scale=.5]{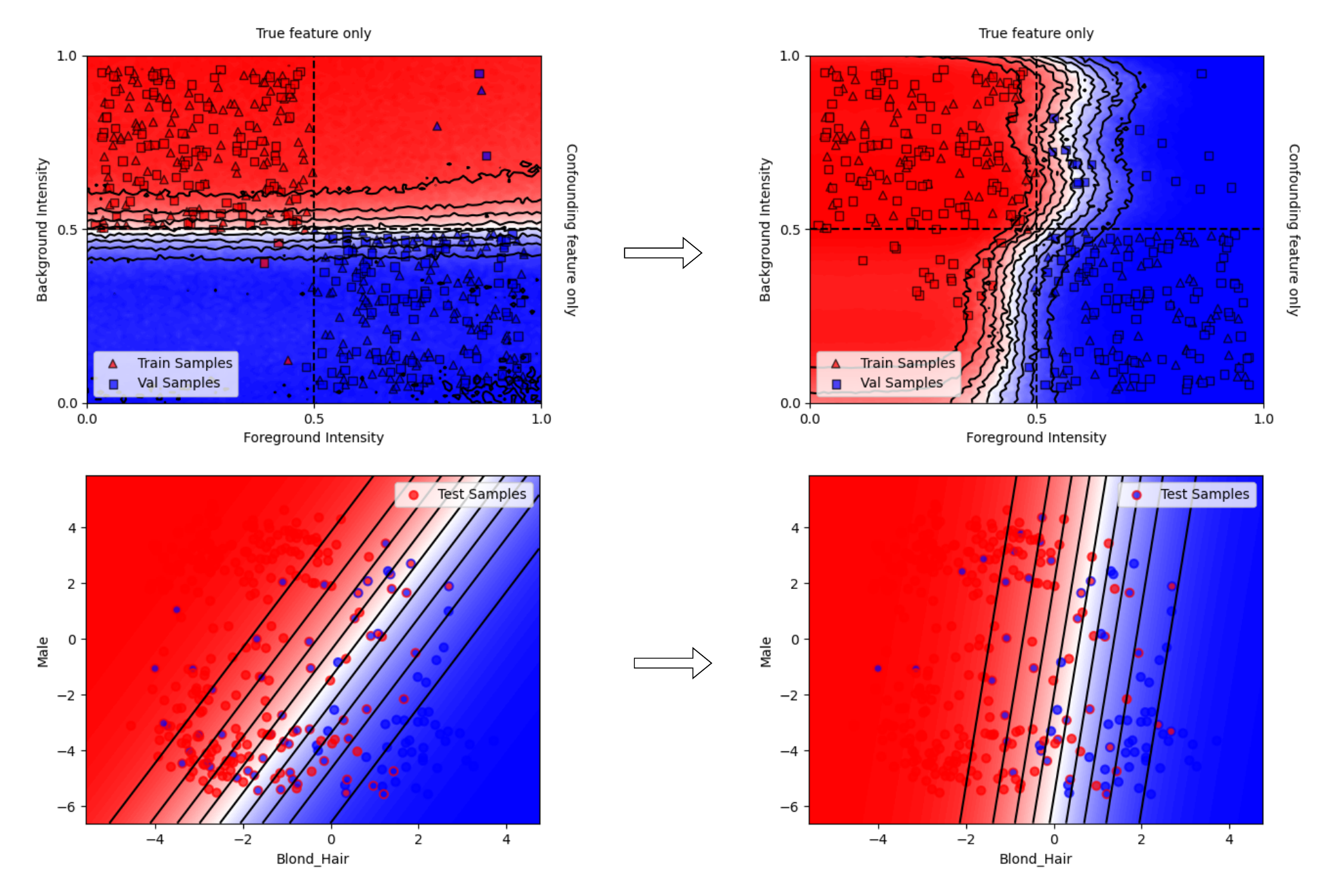}
\caption{Decision boundary of the ResNet-18 student before and after DiDAE-CFKD. The x-axis is the causal feature, the y-axis the confounding one. For Square the boundary is exact, since the whole dataset can be sampled and the x- and y-positions marginalized out; for CelebA it is approximated by projecting onto the oracle's \texttt{Male} and \texttt{Blond\_Hair} predictions (Appendix~\ref{section:desiderata}). Before CFKD the student relies almost entirely on the background on Square and about equally on both factors on CelebA; after one CFKD iteration it relies mostly on the causal feature in both cases.}
\label{fig:didae_cfkd_before_after}
\end{figure*}

\fi

\end{document}